\documentclass[conference]{IEEEtran}
\IEEEoverridecommandlockouts
\usepackage[pagebackref,breaklinks,colorlinks]{hyperref}
\usepackage{cite}
\usepackage{amsmath,amssymb,amsfonts}
\usepackage{graphicx}
\usepackage{textcomp}
\usepackage{caption}
\usepackage{url}
\usepackage{cleveref}
\usepackage{multirow}
\usepackage{algpseudocode}
\usepackage{algorithm}
\usepackage{xcolor} 
\usepackage{colortbl}
\usepackage{pifont}
\usepackage{multirow}
\usepackage{dblfloatfix}

\makeatletter
\renewcommand{\paragraph}{%
  \@startsection{paragraph}{4}%
  {\z@}{0.2em}{-1em}%
  {\normalfont\normalsize\bfseries}%
}
\makeatother

\newcommand{\xmark}{\ding{55}}%
\newcommand{\atan}{\operatorname{atan2}}
\newcommand{\norm}[1]{\left\lVert#1\right\rVert}

\definecolor{red}{rgb}{0.95,0.4,0.4}
\definecolor{green}{rgb}{0.55,1.0,0.55}
\definecolor{lightgreen}{rgb}{0.75,1.0,0.75}
\definecolor{blue}{rgb}{0.4,0.4,0.95}
\definecolor{darkblue}{rgb}{0,0,0.8}
\definecolor{darkred}{rgb}{0.8,0,0}
\definecolor{darkgreen}{rgb}{0,0.5,0}
\definecolor{grey}{rgb}{0.6,0.6,0.6}
\definecolor{amber}{RGB}{255,210,43}
\definecolor{lightgray}{rgb}{0.97,0.97,0.97}

\def\BibTeX{{\rm B\kern-.05em{\sc i\kern-.025em b}\kern-.08em
    T\kern-.1667em\lower.7ex\hbox{E}\kern-.125emX}}
\begin{document}
\pagestyle{plain} 
\pagenumbering{arabic} 

\title{MonoSOWA: Scalable monocular 3D Object detector Without human Annotations
}

\author{\IEEEauthorblockN{Jan Skvrna}
\IEEEauthorblockA{\textit{Dept. of Cybernetics} \\
\textit{FEE, Czech Technical University}\\
skvrnjan@fel.cvut.cz}
\and
\IEEEauthorblockN{Lukas Neumann}
\IEEEauthorblockA{\textit{Dept. of Cybernetics} \\
\textit{FEE, Czech Technical University}\\
lukas.neumann@cvut.cz}
}

\setlength{\abovedisplayskip}{8pt}
\setlength{\belowdisplayskip}{8pt}

\twocolumn[{%
\renewcommand\twocolumn[1][]{#1}%
\maketitle
\begin{center}
    \centering
    \captionsetup{type=figure}
    \begin{tabular}{c}
    \includegraphics[width=0.8\linewidth]{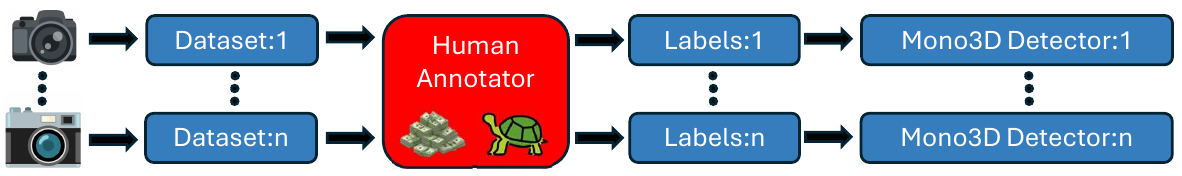} \\  
    \small(a) Traditional pipelines require human labelling and produce new model for every different camera setup\vspace{3pt} \\
    \includegraphics[width=0.95\linewidth]{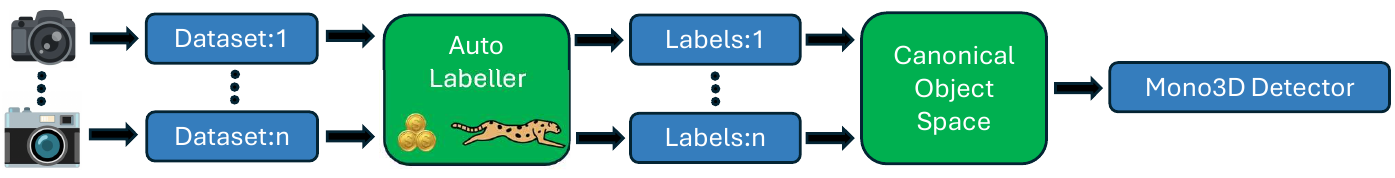}\\  
    \small(b) Our method does not need human labelling and combines multiple camera setups into a single model operating in canonical object space \\
\end{tabular}
    \captionof{figure}{Traditional pipelines (a) vs. the proposed pipeline (b) for monocular 3D object detection.}
    \label{img:teaser}
\end{center}%
}]

\begin{abstract}
Inferring object 3D position and orientation from a single RGB camera is a foundational task in computer vision with many important applications. Traditionally, 3D object detection methods are trained in a fully-supervised setup, requiring LiDAR and vast amounts of human annotations, which are laborious, costly, and do not scale well with the ever-increasing amounts of data being captured.

We present a novel method to train a 3D object detector from a single RGB camera without domain-specific human annotations, making orders of magnitude more data available for training. The method uses newly proposed Local Object Motion Model to disentangle object movement source between subsequent frames, is approximately 700 times faster than previous work and compensates camera focal length differences to aggregate multiple datasets.

The method is evaluated on three public datasets, where despite using no human labels, it outperforms prior work by a significant margin. It also shows its versatility as a pre-training tool for fully-supervised training and shows that combining pseudo-labels from multiple datasets can achieve comparable accuracy to using human labels from a single dataset. The source code and model are available at \url{https://github.com/jskvrna/MonoSOWA}.

\end{abstract}    
\section{Introduction}
\label{sec:intro}

Monocular 3D object detection is a key component of many important systems, ranging from robotics to autonomous cars. 3D object detection methods are typically trained in a fully-supervised setup, requiring specialized vehicles to simultaneously capture RGB image and LiDAR modalities, followed by human annotators manually creating 3D labels. This is extremely laborious, costly, and limits the diversity of training data being captured. To make matters even worse, especially in autonomous driving, new camera setups emerge constantly, which typically means this laborious and costly process needs to be repeated again.

In this paper, we address both of these challenges -- costly human labelling and reliance on dedicated vehicles to capture data -- by introducing a new method to \textit{train a monocular 3D rigid object detector without domain-specific human labels and without LiDAR}. Our method allows us to exploit large amounts of training data readily available because it does not rely on human annotations, which traditionally is the main limiting factor in exploiting captured data, and it does not need additional sensors such as LiDAR, which implies our method can directly exploit data collected by a majority of currently manufactured vehicles, which typically only have a single camera. This is in contrast with some methods which require LiDAR (and human annotations)~\cite{peng2021weakm3d, zakharov2020autolabeling} as specialized vehicles are needed to collect the data. This then severely limits the diversity of geographies, environments, and traffic situations captured.

In our method, we automatically create 3D object labels (7-DOF poses) by using outputs of two generic off-the-shelf models -- a 2D object detector and a monocular depth estimator -- and by combining them into a single pipeline that processes video sequences captured from a moving vehicle. The proposed pipeline uses temporal consistency of objects in real world to precisely infer object 3D object orientation, position and dimensions in a series of consecutive steps, using newly proposed Local Object Motion Model (LOMM). LOMM disentangles the position change between subsequent frames into ego-motion of the vehicle capturing the scene and motion of the labeled object itself, allowing our method, unlike previous work~\cite{tao2023weakly, liu2024vsrd}, to aggregate information for both stationary and moving objects throughout the whole video sequence, which then results in more precise 3D orientation and position 3D label in every frame. To address the heterogeneity of captured data in practice, we are the first weakly-supervised method to address this issue by adopting Canonical Object Space (COS)~\cite{yin2023metric3d, brazil2023omni3d} to effectively harvest and auto-label video sequences captured by different camera setups. 

To summarize, we make the following contributions:
\begin{enumerate}
  \item To the best of our knowledge, our method is the first method to train a monocular 3D object detector for autonomous driving \textit{without 2D or 3D human annotations}; we demonstrate to the community that such an approach is feasible and is worth pursuing further.    
    \item We propose a novel auto-labelling pipeline which exploits temporal consistency in video sequences to generate 3D position, size, and orientation labels of rigid objects (cars) from a monocular camera. The pipeline uses novel Local Object Motion Model (LOMM) to disentangle ego-motion of the vehicle capturing the scene and motion of the labelled object itself to aggregate information both for stationary and moving objects.
    \item Our auto-labelling method is \textit{almost 700 times faster} than previous work and is able to aggregate data captured by different camera setups, allowing us to process significantly more data than the current state-of-the-art methods~\cite{liu2024vsrd}, where their auto-labelling speed in practice prohibits applications to large-scale datasets.
    
\end{enumerate}

\section{Related work}

\paragraph{Monocular Depth Estimation.}
Monocular depth estimation has shown significant improvements in recent years. MiDAS~\cite{ranftl2020towards} employs affine-invariant losses to enable training on multiple datasets. ZoeDepth~\cite{bhat2023zoedepth} combines training on multiple datasets with relative losses and further fine-tunes on the target metric dataset. Depth Anything~\cite{yang2024depth} focuses on the affine-invariant depth estimation using self-supervised pre-training. For metric depth estimation, which is relevant for 3D object detection, it fine-tunes on the target dataset such as KITTI~\cite{geiger2013vision}. Metric3D~\cite{hu2024metric3d, yin2023metric3d} focuses on \textbf{metric} depth estimation as it uses a Canonical Camera Transformation Module to further improve zero-shot performance, together with DINOv2~\cite{oquab2023dinov2} or ConvNext~\cite{liu2022convnet} as the backbone and DPT~\cite{ranftl2021vision} as the predictor. We opt to use Metric3D~\cite{hu2024metric3d, yin2023metric3d} for its superior zero-shot performance.

\paragraph{Fully-supervised Monocular 3D Object Detection.}
Early methods~\cite{weng2019monocular, wang2019pseudo, ma2019accurate} were based on lifting the image into pseudo-LiDAR and then detecting objects in the generated point cloud. SMOKE~\cite{liu2020smoke}, on the other hand, directly predicts 3D bounding boxes from images. MonoDTR~\cite{huang2022monodtr} and MonoDETR~\cite{zhang2023monodetr} use an end-to-end depth-aware/guided transformer~\cite{vaswani2017attention} architecture to directly predict 3D bounding boxes. SSD-MonoDETR~\cite{he2023ssd} introduces scale awareness into the model. MonoATT~\cite{zhou2023monoatt} variably assigns tokens to areas of higher importance. MonoFlex~\cite{zhang2021objects} uses an ensemble of direct depth and keypoint-based predictions and decouples the detection of truncated and untruncated vehicles. MonoCD~\cite{yan2024monocd} enhances the ensemble with a complementary depth predictor. However, all the aforementioned methods were previously trained using 3D human labels. We utilize MonoDETR~\cite{zhang2023monodetr} for its competitive performance and high-quality code implementation.

\paragraph{Weakly-supervised LiDAR 3D Object Detection.}
The majority of previous work on weakly-supervised 3D detection have focused on training 3D detectors for LiDAR. VS3D~\cite{qin2020weakly} pioneered in this area by using point density in LiDAR scans to identify objects. Zakharov et al.~\cite{zakharov2020autolabeling} employs a 2D off-the-shelf detector accompanied by a novel Signed Distance Fields (SDF)-renderer and Normalized Object Coordinate Space (NOCS). WS3D~\cite{meng2020ws3d, meng2021towards} employs Bird's Eye View (BEV) human-centred clicks as weak supervision to train a network for detecting vehicles. McCraith et al.~\cite{mccraith2022} employ direct fitting of generic templates into LiDAR scans while requiring consistency between frames for stationary vehicles and discounting outliers. TCC-det~\cite{skvrna2025tcc} uses direct fitting of generic templates on aggregated information corresponding to each instance and trains with two additional losses for finer alignment. VG-W3D~\cite{huang2024vgw3d} encodes the image and LiDAR scan separately and then enforces alignment between 2D and 3D on multiple feature levels to predict precise 3D boxes.
\begin{figure*}
\centering
\includegraphics[width=\textwidth]{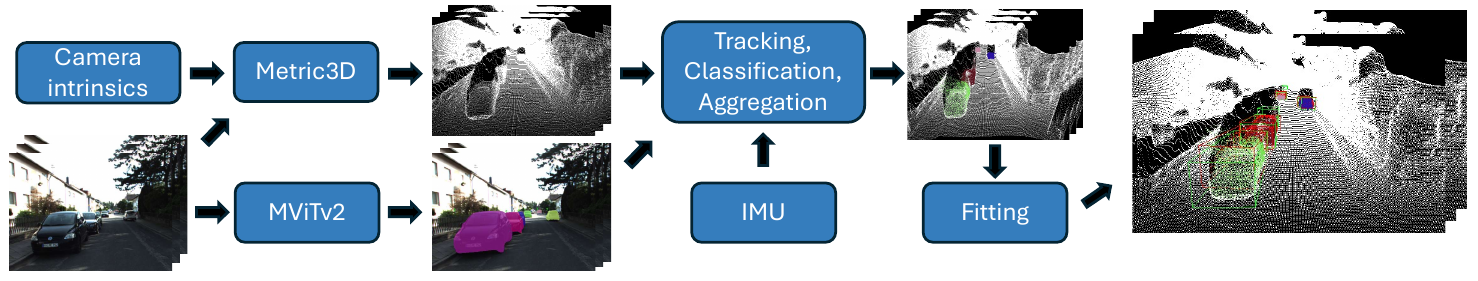}
\caption{Our auto-labelling pipeline creates 3D labels using only images and ego-motion data as inputs (and known camera parameters).}
\label{fig:autolabel_pipeline}
\end{figure*}

\paragraph{Unsupervised LiDAR 3D Object Detection}

Several approaches have been proposed for unsupervised object detection and tracking in LiDAR scans. Cheng et al.~\cite{cen2021open} focus on open-set 3D object detection using the MLUC network, which identifies objects while keeping some unclassified. ClusterNet~\cite{wang20224d} performs unsupervised object discovery using both 3D LiDAR and 2D RGB images. Najibi et al. ~\cite{najibi2022motion} involve scene-flow estimation and meta-label creation for detection and trajectory prediction. MODEST~\cite{you2022learning} detects objects via clustering and refines them in a self-training cycle. DRIFT~\cite{luo2023reward} uses a Persistency prior score for noisy label creation and reinforced learning for fine-tuning. Recent approaches include density-based clustering and self-supervised learning for object detection~\cite{zhang2023towards}, clustering based on the "what moves together" paradigm~\cite{seidenschwarz2024semoli}, and self-supervised LiDAR flow estimation for motion cue generation~\cite{baur2024liso}. While \cite{cen2021open, zhang2023towards} do not differentiate between moving and stationary vehicles, \cite{wang20224d, najibi2022motion, you2022learning, luo2023reward, seidenschwarz2024semoli, baur2024liso} only detect  moving vehicles and assume the detector will generalize to detect all movable vehicles (including stationary).

\paragraph{Weakly-supervised Monocular 3D Object Detection}
WeakMono3D~\cite{tao2023weakly} is the first method not using LiDAR scans, as only sequences of images with 2D ground-truth labels are required as input.  VSRD~\cite{liu2024vsrd} represents each instance as a surface in Signed Distance Fields (SDF). The SDFs are rendered into masks through their proposed volumetric-aware silhouette rendering, enabling end-to-end training. Each instance is optimized over multiple frames. Both WeakMono3D and VSRD struggle with moving vehicles, which are either discarded or given low confidence; on the other hand, our work can leverage information from both moving and stationary vehicles. While WeakMono3D and VSRD show promising results in both weakly-supervised and semi-supervised training, they require 2D instance masks, and in VSRD rendering of each frame takes approximately 15 minutes. In contrast, our method \emph{does not require any human annotations at all} and is approximately 700 times faster. 
\section{Method}
\label{sec:method}

The process of automatic label creation assumes that images are captured in a sequence, which is a natural way to capture data in driving scenarios. It also assumes that the camera setup is known (both intrinsics and extrinsics) for each sequence, although each sequence might have different camera setup. Lastly, it assumes approximate data about ego-vehicle's motion is available (e.g. from GPS or IMU), which again is commonplace. Because labelling a 7-DOF pose of objects jointly (position, size, yaw) is a very challenging problem, we instead infer individual pose parameters separately in a sequence of steps (see \cref{fig:autolabel_pipeline}).

\subsection{Pseudo-LiDAR aggregation}
The auto-labelling process begins by inferring a metric depth map $D\in \mathbb{R} ^ {h \times w}$ for each camera image $I \in \mathbb{R}^{3 \times h \times w}$, where $h$ and $w$ denotes image height and width, using an off-the-shelf monocular depth estimator. We opted to use Metric3D~\cite{yin2023metric3d}, as it is zero-shot, it does not limit our method in terms of data distribution and it has shown great generalization results. 

Using the inferred depth map $D$, a 3D point cloud is generated as
\begin{align}    
    X_{u, v} &= \frac{D_{u, v} \cdot (u - c_x)}{f_x} \qquad Y_{u, v} = \frac{D_{u, v} \cdot (v - c_y)}{f_y} \nonumber \\
    \label{equation:pcloud_generation} Z_{u, v} &= D_{u, v}
\end{align}
where $u \in [0, h]$ and $v \in [0, w]$, $c$ is the principal point and $f$ is the focal length. We denote the pseudo-LiDAR point cloud as $P_i \in \mathbb{R}^{3\times hw}$, where $i$ stands for the frame index. 

\begin{figure}
\centering
\renewcommand{\arraystretch}{0.3} 
\begin{tabular}{c}
    \includegraphics[width=0.8\linewidth]{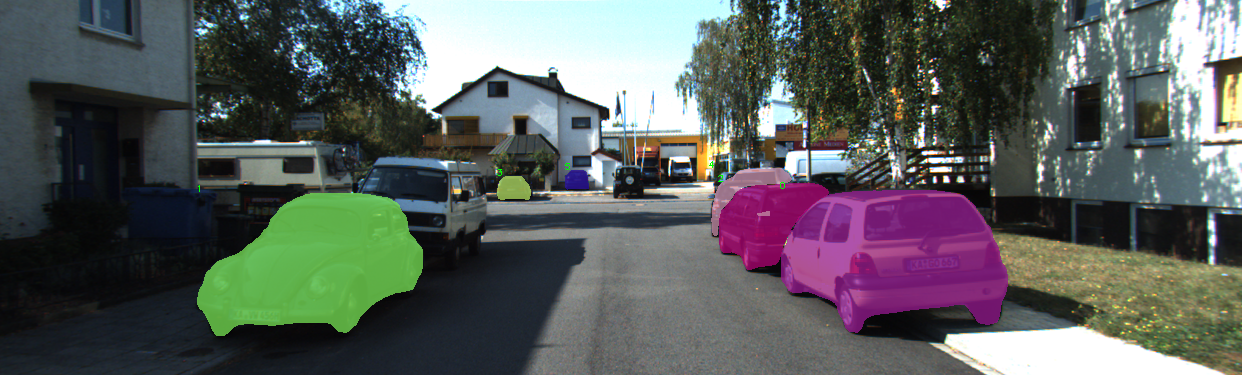} \\
    \includegraphics[width=0.8\linewidth]{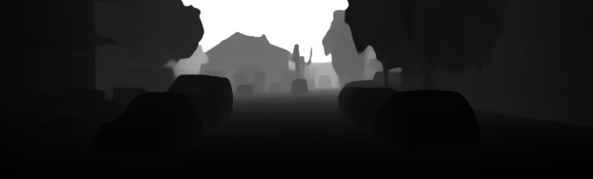} \\
    \includegraphics[width=0.8\linewidth, trim=0 0 300 300, clip]{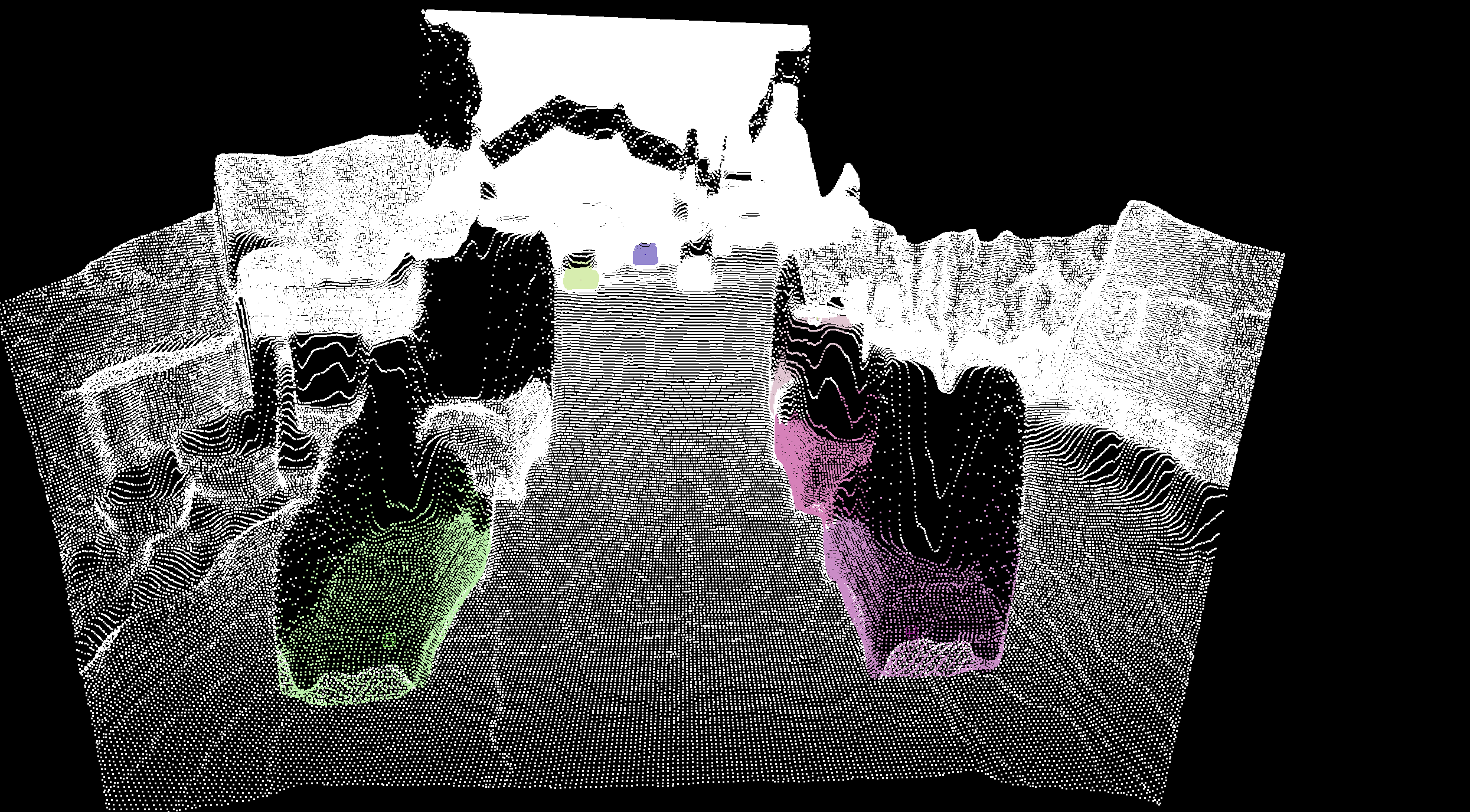}
\end{tabular}
\captionof{figure}{Combining instance segmentation in image (top) with monocular metric depth estimation (middle) enables us to produce pseudo-LiDAR point cloud for each object (bottom).}
\label{fig:pseudo_lidar_mask}
\end{figure}

Next, inspired by~\cite{mccraith2022}, we employ an off-the-shelf 2D object image detector~\cite{li2022mvitv2} to detect the objects we are interested in (vehicles) and their instance segmentation masks in each camera image. Given an instance segmentation mask $M_{i,j} \in \mathbb{R}^{h\times w}$ of an object $j$ in a frame $i$, we find object point cloud $P_{i,j}$ in the point cloud as
\begin{align}
    P_{i,j} = \big\{ p \in P_i\,|\, \mathcal{T}(p) \in M_{i,j} \big\}
\end{align}
where $\mathcal{T}(p)$ denotes the projection of the 3D point $p$ into image coordinates, as shown in \cref{fig:pseudo_lidar_mask}. 

\paragraph{Tracking.} In order to exploit temporal consistency of given object across the sequence frames, we need first to establish correspondences of the object instance between frames. Because we have inferred the approximate 3D point cloud of each object instance $P_{i,j}$ in each frame, we track each object in the 3D world coordinate system. 

We approximate spatial location $L_{i,j} \in \mathbb{R}^3$ of each object $j$ in frame $i$ as the median of its point cloud $P_{i,j}$. The tracking is initialized in $n$ frames before the reference (current) frame, and for each frame, it matches the instances based on locations in the current frame $L_{i,j}$ and the predicted future locations from the last frame into the current frame $\hat{L}_{i,j}$. For the location prediction $\hat{L}_{i,j}$, a simple physics-aware motion model is used 
\begin{align}   
    \label{equation:motion_model}
    \hat{L}_{i,j} &= L_{i-1, j} + \frac{1}{3}\sum_{k=1}^3 L_{i-k, j} - L_{i-k-1, j} 
\end{align}
To match two object instances between frames, it is required that the instances are both their nearest neighbours and that the distance between them is lower than a set threshold $T_{\text{dist}}$. Otherwise, the instance is considered as a new object. After this step, each object instance is represented by extracted points $P_{i,j}$ and locations $L_{i,j}$ in each frame the instance is present. The tracking sequentially processes all the frames up to $m$ after the reference frame. Finally, we transform all $P_{i,j}$ into the reference frame coordinate space (current frame for which labels are generated) using vehicle ego-motion data.

\subsection{Local Object Motion Model}
In our method, we classify object instances as either stationary or moving, as different principles of temporal consistency exploitation are applied to those classes. Note that all object instances actually appear to change position between frames -- they change their location in relative terms, as the ego-vehicle itself is driving while data is captured. 

In order to classify a vehicle as stationary or moving, we propose the following novel classification procedure.

For each instance, all locations $L_{m...n, j}$ are taken, and differences $\Delta_{i,j}$ between all adjacent locations are computed, where $k$ is the first frame, where the instance is present and $l$ is the last. From those differences, the standard deviation $\sigma$ is calculated as
\begin{align}
    \label{equation:sigma}    
    \sigma_{j} &= \frac{1}{\sqrt{2}} \sqrt \frac{1}{l-k}\sum_{i=k+1}^l (\mu_j - \Delta_{i,j})^2 \\
    \mu_j &= \frac{1}{l-k}\sum_{i=k+1}^l \Delta_{i,j} \qquad \Delta_{i,j} = L_{i,j} - L_{i-1,j}
\end{align}

Furthermore, to decide if the difference between the $\norm{\mu_j}_2$ and $\norm{\sigma_j}_2$ is statistically significant or not, the ratio $z$ is calculated as
\begin{equation}
    \label{equation:zscore}
    z_j = \frac{\norm{\mu_j}_2}{\norm{\sigma_j}_2}
\end{equation}

To classify the instance as moving, the ratio $z$ is required to be higher than the threshold $T_{z}$ and simultaneously, the net distance must be higher than a given threshold $T_{m}$. This follows the idea, that if the car is stationary, the $\norm{\mu_j}_2$ will be significantly lower than $\norm{\sigma_j}_2$ as the movement of the vehicle is introduced by jitter from pseudo-LiDAR. On the other hand, if the vehicle is moving the $\norm{\mu_j}_2$ will be significantly higher than $\norm{\sigma_j}_2$.

As a result, for moving instances, we estimate the trajectory in adjacent frames. Given the known trajectory and the fact that cars move by physics constraints, estimating the yaw (the orientation) of the car is trivial. This simple fact simplifies the autolabelling significantly.

\begin{figure}
\centering
\setlength{\tabcolsep}{1pt}
\begin{tabular}{c c}
    \includegraphics[width=0.45\linewidth]{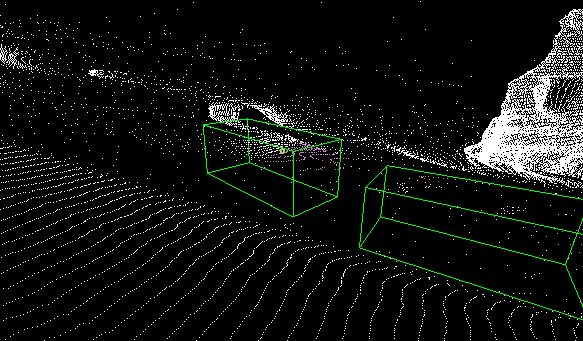} & \includegraphics[width=0.45\linewidth, trim=0 6 0 6, clip]{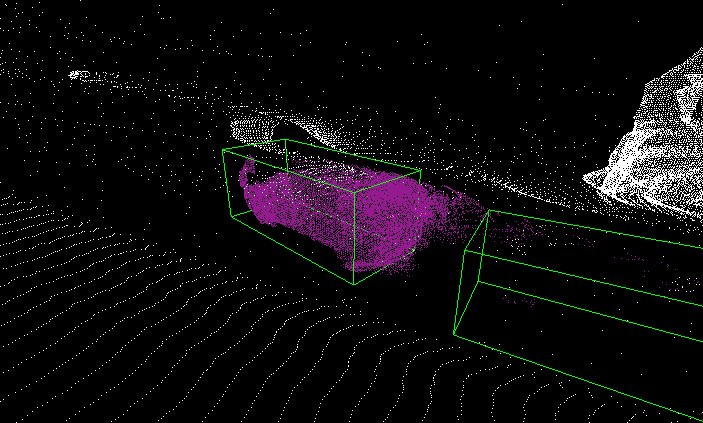}\\ 
    \small{a)} & \small{b)} \\
    \includegraphics[width=0.45\linewidth,  trim=0 18 0 18, clip]{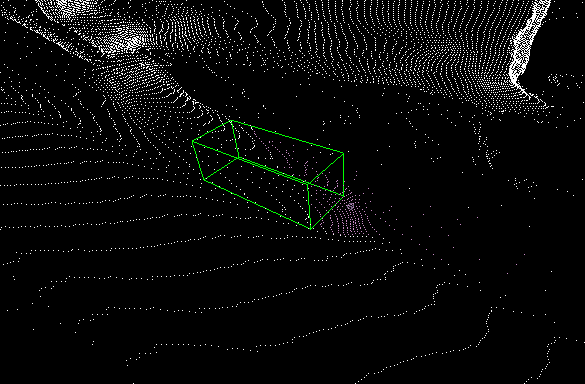} & \includegraphics[width=0.45\linewidth, trim=0 0 0 0, clip]{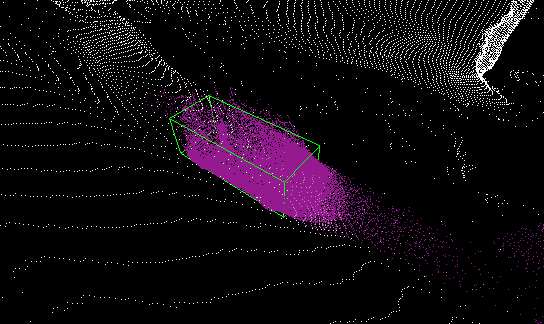}\\ 
    \small{c)} & \small{d)}\\
\end{tabular}
\caption{The point cloud of a significantly occluded vehicle (a) is recovered in subsequent frames (b).  Distant car with poor depth prediction (c) is refined as the ego-vehicle drives closer (d). Pseudo-LiDAR of each frame in white, aggregated object points in purple, human label in green for reference.}
\label{fig:aggregated_points}
\end{figure}

On the other hand, for the stationary instances, we can simply aggregate points $A_{j}$ by concatenating all $P_{i,j}$, as all $P_{i,j}$ are in the reference frame. This creates a much denser and more informative representation of an instance, and it can also recover from some mistakes, as shown in \cref{fig:aggregated_points}.

\subsection{Step-wise autolabelling}

Finding oriented 3D bounding boxes is a 7-degrees of freedom (DOF) problem. Instead of solving it directly, our method divides it into three disjunctive problems: orientation, location and size estimation.

\paragraph{Orientation and size estimation.} Orientation estimation is different for moving and stationary instances. For moving instances, the yaw of the object is estimated by directly computing the angle between adjacent locations as
\begin{equation}
    \label{equation:angle_est}
    \theta_{i, j} = \atan\left(\frac{L_{i, j}(z) - L_{i-1, j}(z)}{L_{i, j}(x) - L_{i-1, j}(x)}\right)
\end{equation}
To make it more robust, yaws for up to 5 locations before and after the reference frame are predicted, and then the median is taken out of those yaw predictions.

For stationary instances, we improve the 3D box fitting algorithm of Zhang et al.~\cite{zhang2017efficient}. The algorithm first flattens all points into Bird's eye view (BEV), and then it iterates over all possible angles $\theta \in {0, \frac{\pi}{2}}$. For each angle, two perpendicular axes to each other are computed, and all points are projected into both axes $x \in \mathbb{R}$ and $y \in \mathbb{R}$.

\begin{algorithm}
\footnotesize
\caption{\small Saturated Closeness Criterion}
\label{algo:saturated}
\begin{algorithmic}[1]
\Require \( \alpha \) \Comment{Steepness parameter}
\Function{SaturatedCloseness}{$x, y$}
    \State \( x^{\text{max}} \gets P_{90}(x) \), \( x^{\text{min}} \gets P_{10}(x) \) \Comment{P is percentile}
    \State \( y^{\text{max}} \gets P_{90}(y) \), \( y^{\text{min}} \gets P_{10}(y) \)
    \State \( x \gets \arg\min_{v \in \{x^{\text{max}} - x, x - x^{\text{min}}\}} \| v \|_2 \)
    \State \( y \gets \arg\min_{v \in \{y^{\text{max}} - y, y - y^{\text{min}}\}} \| v \|_2 \)
    \State \( x \gets \sigma(\alpha \cdot x)\) \Comment{$\sigma$ is the sigmoid function}
    \State \( y \gets \sigma(\alpha \cdot y)\)    
    \State \( \mathcal{L} \gets 0 \)
    \For{ \( i = 1 \) to \( \text{length}(x) \) }
        \State \( e \gets \min(x(i), y(i)) \)
        \State \( \mathcal{L} \gets \mathcal{L} + e\)
    \EndFor
    \State \Return \( \mathcal{L} \)
\EndFunction
\end{algorithmic}
\end{algorithm}

For each $\theta$, we calculate a newly proposed Saturated Closeness Criterion (\cref{algo:saturated}). Instead of taking a simple sum of distances $x$ and $y$, as Closeness Criterion in Zhang et al.~\cite{zhang2017efficient}, we feed the $x$ and $y$ into a sigmoid multiplied by steepness parameter $\alpha$ (Line 6 and 7 of \cref{algo:saturated}), inspired by Template Fitting Loss in \cite{skvrna2025tcc}, to saturate the distance of outliers and reduce their influence on the fitting. As a result, the robustness and accuracy of estimation in noisy point clouds are significantly improved over the original algorithm.

Another modification is that Closeness Criterion computes the distance to either minimum or maximum. However, those extremal points tend to be outliers. To mitigate the problem, each point is assigned either to the 10th or 90th percentile of the projected points instead of min/max. The assignment is based on the minimal distance to one of the extremal points. Intuitively, this creates an axis at the most probable location, mitigating outliers' influence. 
The algorithm chooses the yaw with a minimal criterion value and creates two hypotheses, as it doesn't differentiate between the front and back of the car. 

The algorithm also outputs a spatial dimensions estimate for a single frame. If any of the values exceed the typical dimensions of an instance, the output is replaced by a prior estimate of the dimensions of a generic instance. Also, it is necessary to take into account that estimating the spatial dimensions of an instance from a single image can be an ill-posed problem; for example, if an instance is seen from the back, it is not possible to estimate length correctly. Thus, such cases are detected by computing the difference between car orientation and the viewing angle of the car and if the difference is ${0, \frac{\pi}{2}, \pi, \frac{3\pi}{2}}$, then the output is replaced by a prior car dimensions estimate.

\paragraph{Position refinement.}
The yaw $\theta$ and a prior 3D position are already known from the previous steps. In order to get a more fine-grained estimate of position, we apply small perturbations (up to 2 meters) along $x$ and $z$ axes and use Template Fitting Loss (TFL)~\cite{skvrna2025tcc} as the criterion to select the final position. To compute TFL, we use a generic template mesh of a vehicle sampled into a point cloud. Given a generic vehicle point cloud, we move the point cloud according to perturbations and compute TFL at those points. The lowest TFL value corresponds to the best fit between those point clouds. Unlike Chamfer Distance, it works well with outliers. In this stage, we also determine whether the car is facing towards or away from the ego-vehicle by testing two hypotheses for vehicle orientation  -- $\theta$ and $\theta + \pi$ and again selecting the orientation with the lowest TFL.

\begin{figure}
\centering
\includegraphics[width=1\linewidth]{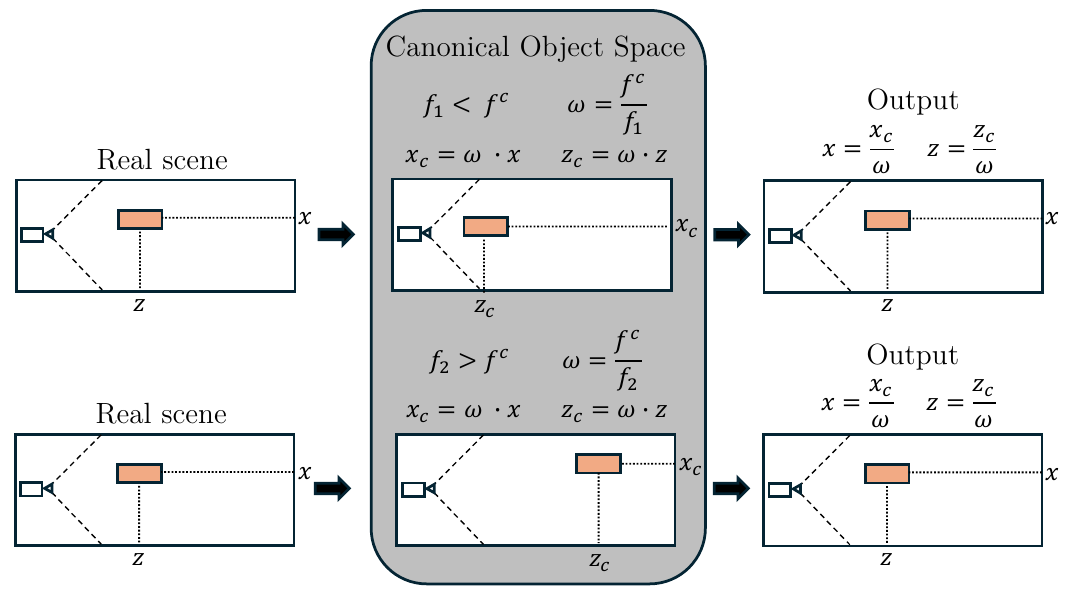}
\captionof{figure}{Generated 3D labels are transformed into Canonical Object Space to accommodate for different focal lengths; the resulting model is therefore trained to output canonical coordinates. At inference time, model predictions are transformed back to world coordinates, using (known) camera parameters.}
\label{fig:canonical_object_space}
\end{figure}
\begin{table*}[h]
\centering
\footnotesize
\setlength{\tabcolsep}{5pt}
\begin{tabular}{l|c|cc|cc|cc|c}
\hline
 & \multicolumn{3}{c|}{\textbf{Training supervision}} & \multicolumn{4}{c|}{\textbf{3D Object detection accuracy}} & \textbf{Labelling} \\
& & \multicolumn{2}{c|}{\textit{Human Labels}} &  \multicolumn{2}{c|}{\textbf{AP\textsubscript{BEV}/AP\textsubscript{3D}@0.5}} & \multicolumn{2}{c|}{\textbf{AP\textsubscript{BEV}/AP\textsubscript{3D}@0.3}} & \textbf{speed} \\
\textbf{Method} & LiDAR & Masks & 3D boxes & Easy & Hard & Easy & Hard & per frame \\
\hline
\multicolumn{8}{c}{\cellcolor{lightgray} Fully-supervised methods} \\ \hline
MonoFlex~\cite{zhang2021objects} & \color{orange}yes*  &   \color{darkgreen}\xmark & \color{red}yes  & 50.82/43.11 & 41.78/34.43 & 69.70/67.07 & 59.86/57.26 & - \\
MonoDETR~\cite{zhang2023monodetr} & \color{orange}yes* &   \color{darkgreen}\xmark  & \color{red}yes  & 47.21/41.01 & 36.05/30.38 & 63.07/60.49 & 54.04/50.03 & - \\ \hline
\multicolumn{8}{c}{\cellcolor{lightgray} Methods requiring some human annotations} \\ \hline
WeakM3D~\cite{peng2021weakm3d} & \color{orange}yes & \color{red}yes &   \color{darkgreen}\xmark  & 8.10/2.96 & 2.96/2.01 & 29.89/21.25 & 24.01/15.34 & 80 ms \\
Autolabels~\cite{zakharov2020autolabeling} & \color{orange}yes & \color{red}yes &   \color{darkgreen}\xmark  & 20.18/4.69 & 14.33/2.79 & 48.16/12.92 & 37.34/9.94 & 6 sec\\
VSRD~\cite{liu2024vsrd} &  \color{darkgreen}\xmark  & \color{red}yes &   \color{darkgreen}\xmark  & 29.07/21.77 & 22.83/16.46 &58.40/50.86 & 50.61/43.45 & 15 min\\ \hline
\multicolumn{8}{c}{\cellcolor{lightgray} Methods without human annotations} \\ \hline
\emph{MonoSOWA (Ours)} &  \color{darkgreen}\xmark  &  \color{darkgreen}\xmark  &  \color{darkgreen}\xmark  & 38.41/29.98 & 35.26/27.56 & 50.84/42.72 & 49.22/46.59 & 1311 ms \\\hline
\end{tabular}
\caption{Evaluation results of monocular 3D object (vehicle) detection on the KITTI-360~\cite{liao2022kitti} test set. * denotes that given data modality was not used by the method itself, but it was required in the annotation process to create human 3D box annotations.}
\label{tab:kitti360_test}
\end{table*}

\subsection{Canonical Object Space (COS)}

Camera focal length is a crucial parameter for precise 3D position estimation, but monocular 3D detectors have no direct way to compensate for changing focal length because the same object at the same distance from the camera will have different size in the image, depending on the focal length.

Taken to the extreme, for the exact same object image measured in pixels, the network needs to predict arbitrarily different distances in world coordinates, depending on camera focal length. This phenomenon confuses the network during training, and it is therefore necessary to solve this problem in order to be able to train a network on data captured by different cameras.

Inspired by methods such as Metric3D~\cite{yin2023metric3d} and Omni3D~\cite{brazil2023omni3d}, our auto-labelling method uses Canonical Object Space (COS) where \textit{3D bounding boxes} are trained to be invariant to changes in the focal length by choosing a single canonical focal length and then transforming the generated labels (objects) into COS. Instead of transforming whole images or depth maps as in Metric3D, only the $x,y,z$ coordinates of pseudo-ground truth labels are transformed directly when creating the pseudo-labels. To transform labels, the scaling parameter $\omega_i$ is calculated as 
\begin{align}
    \omega_i = \frac{f^C}{f_i},
    \label{method:scaling_parameter}
\end{align}
where $f_i$ stands for the focal length of the frame $i$ and $f^C$ is the canonical focal length. Given the scaling parameter $\omega_i$, the object instance $(x,y,z)$ is transformed into COS as
\begin{equation}
    x^C = x\cdot \omega_i \qquad y^C = y \cdot \omega_i \qquad z^C = y \cdot \omega_i.
    \label{method:canonical_depth}
\end{equation}
The model then operates in the COS and is directly supervised by the transformed labels (objects). During inference, all predictions are transformed from COS using the scaling parameter $\omega_j$ of the inference frame $j$ (see \cref{fig:canonical_object_space}).

This very simple transformation enables the network to train on multiple datasets where focal length differs and also allows the network to \textit{work in unseen camera setup} (focal length) because $\omega_j$ is calculated during inference.

Note that in the training process, it is necessary to adapt for data augmentations that affect the perceived focal length of the image, such as image scaling, and for the fact that the detector resizes the image into a constant input size by adjusting the perceived focal length of the image subject to the applied augmentations and the resizing.

\section{Experiments}

\begin{table}
\centering
\footnotesize
\setlength{\tabcolsep}{3pt}
\begin{tabular}{l c | c c}

 & \multicolumn{2}{c}{\textbf{AP\textsubscript{BEV}/AP\textsubscript{3D}@0.5}}\\
\textbf{Method} & \textbf{Published} & Easy & Hard \\
\hline
Autolabels~\cite{zakharov2020autolabeling}& ICCV2021& \underline{51.85}/4.65 & \underline{46.10}/2.92 \\
VSRD~\cite{liu2024vsrd} & CVPR2024 & 47.12/\underline{35.25} & 43.91/\underline{32.64}\\
Ours & - & \textbf{61.17}/\textbf{47.07} & \textbf{51.92}/\textbf{45.51}
\end{tabular}
\caption{Auto-labelling AP on KITTI-360~\cite{liao2022kitti} training set.}
\label{tab:pseudo-labels}
\end{table}

\begin{table*}
    \centering
    \footnotesize
    \begin{tabular}{l|c|cc|cccc|cccc}
         \hline

     & \multicolumn{3}{c|}{\textbf{Training supervision}} & \multicolumn{8}{c}{\textbf{3D Object detection accuracy (Level 2)}} \\
& & \multicolumn{2}{c|}{\textit{Human Labels}} &  \multicolumn{4}{c|}{\textbf{AP\textsubscript{BEV}@0.5}} & \multicolumn{4}{c}{\textbf{AP\textsubscript{3D}@0.5}}  \\
\textbf{Method} & LiDAR & Masks & 3D boxes & All & 0m-30m & 30m-50m & 50m-$\infty$ & All & 0m-30m & 30-50m & 50m-$\infty$ \\
         \hline
         DEVIANT~\cite{kumar2022deviant} & \color{orange}yes*  &   \color{darkgreen}\xmark & \color{red}yes & \xmark & \xmark & \xmark & \xmark & 10.29 & \underline{26.75} & 4.95 & 0.16 \\
         MonoDETR~\cite{zhang2023monodetr} & \color{orange}yes*  &   \color{darkgreen}\xmark & \color{red}yes & \textbf{23.63} & \textbf{39.10} & \textbf{20.95} & \textbf{4.70} & \textbf{21.41} & \textbf{37.89} & \textbf{18.61} & \underline{3.69}  \\
         \emph{MonoSOWA (Ours)} &  \color{darkgreen}\xmark  &  \color{darkgreen}\xmark  &  \color{darkgreen}\xmark & \underline{18.98} & \underline{33.51} & \underline{12.02} & \underline{4.55} & \underline{13.46} & 24.65 & \underline{5.87} & \textbf{4.55}  \\
         \hline
    \end{tabular}
    \caption{Monocular 3D object (vehicle) detection AP on Waymo~\cite{Sun_2020_CVPR} validation set.}
    \label{tab:waymo}
\end{table*}

\paragraph{Datasets.} We use three public datasets for evaluation -- KITTI~\cite{geiger2013vision}, KITTI-360~\cite{liao2022kitti} and Waymo~\cite{Sun_2020_CVPR}. 
In the KITTI dataset, we employ the right RGB camera; in the KITTI-360 dataset, we use the right perspective camera; and in Waymo, we utilize the front-facing camera. Note that the camera parameters -- their focal length, resolution, and FOV -- are different in each dataset. In KIITI, we use the same training (3,712 samples) and validation (3,769 samples) splits as~\cite{zakharov2020autolabeling,wei2021fgr,mccraith2022,skvrna2025tcc} are used. In KITTI-360 dataset, training (6 sequences, 44,178 frames) and testing (1 sequence, 2,459 frames) splits from VSRD~\cite{liu2024vsrd} are used. Waymo experiments follow the official training (798 sequences, 158,080 frames) and validation (202 sequences, 39,988 frames) splits. 

\paragraph{Implementation.}
We used off-the-shelf MViT2-Huge~\cite{li2022mvitv2} in the Detectron2 framework~\cite{wu2019detectron2} trained on MS-COCO~\cite{lin2014microsoft}  as the 2D object detector and Metric3DV2-giant~\cite{hu2024metric3d} for metric depth estimation. We use our pseudo-labels to train MonoDETR~\cite{zhang2023monodetr} monocular 3D object detector, using AdamW~\cite{loshchilov2017decoupled} optimiser with learning rate and weight decay equal to 0.0002 and 0.0001, respectively, while keeping other hyper-parameters as in ~\cite{zhang2023monodetr}. The canonical focal length $f_c$ is equal to 750. Aggregation is done over 100 frames, the $T_{z}$ threshold for stationary/moving classification is 0.2, and the minimum threshold $T_m$ is 5 meters. The steepness parameter $\alpha$ in the Saturated Closeness Criterion is 10 (see Supplementary material for ablations). Note that we used the same hyper-parameter settings for all three datasets.

\subsection{Results}

\begin{table}
\centering
\footnotesize
\setlength{\tabcolsep}{3pt}
\begin{tabular}{l c | c c c }
\hline
\textbf{Training}& \textbf{Fine-tuning} & \multicolumn{3}{c}{\textbf{AP\textsubscript{BEV}/AP\textsubscript{3D}@0.7}}\\
\textit{\color{darkgreen} pseudo labels} & \textit{\color{orange} human labels} & Easy & Moderate & Hard \\
\hline
VSRD~\cite{liu2024vsrd} & \multirow{2}{*}{0\%} & 0.002/0.001 & 0.004/0.001 & 0.005/0.002 \\
Ours & & 24.46/8.24 & 20.03/6.84 & 16.14/6.36 \\
\hline
VSRD & \multirow{2}{*}{25\%} & 31.72/21.76 & 22.32/15.43 & 18.86/12.55 \\
Ours & & 39.99/32.64 & 30.92/25.06 & 26.47/20.35 \\
\hline
VSRD & \multirow{2}{*}{50\%} & \underline{43.44}/31.05 & 31.54/21.48 & 27.17/17.93 \\
Ours & & \textbf{44.32}/\underline{33.70} & \textbf{32.97}/\underline{25.65}
& \textbf{27.52/23.08} \\
\hline
VSRD & \multirow{2}{*}{75\%} & 42.58/32.95 & 31.08/24.68 & 27.19/21.38 \\
Ours & & 43.32/\textbf{34.11} & \underline{32.48}/\textbf{25.68} & \underline{27.39}/\underline{23.05} \\
\hline
- & 100\% & 37.99/29.36 & 26.76/20.64 & 23.02/17.30 \\
\hline
\end{tabular}
\caption{Monocular 3D object detection evaluated on KITTI~\cite{geiger2013vision} validation set. Pre-training on KITTI-360~\cite{liao2022kitti} pseudo-labels and fine-tuning on a fraction of the KITTI training set.}
\label{tab:fine-tuning}
\end{table}

\paragraph{KITTI-360.} In the first experiment, we generate pseudo-labels on the KITTI-360~\cite{liao2022kitti} dataset and train MonoDETR~\cite{zhang2023monodetr} with these pseudo-labels. We then compare our trained model with previously reported results of other trained monocular 3D object detectors. As shown in \cref{tab:kitti360_test}, our method \textbf{outperforms current state-of-the-art} weakly-supervised method VSRD~\cite{liu2024vsrd} both in Bird's eye view (BEV) and 3D average precision at 0.5 IoU, despite the fact \textit{our method does not use any human labels}. In the 0.3 IoU metric, our method outperforms both WeakM3D~\cite{peng2021weakm3d} and Autolabels~\cite{zakharov2020autolabeling} by a significant margin both for BEV and 3D and it achieves competitive results with VSRD~\cite{liu2024vsrd}. The remaining gap in the IoU 0.3 metric is caused by KITTI-360 human labels which are amodal (they include occluded or invisible object parts), which significantly helps VSRD that uses them in its training. Our method instead relies on a 2D detector, which only outputs boxes based on what is actually visible in the image, but as shown in \cref{tab:abl_gt_masks}, when identical inputs are used, our method significantly outperforms VSRD in all metrics. Note that our auto-labelling pipeline takes, on average, 1.3 seconds per frame (including all the models it uses as inputs - see \cref{tab:runtime}) -- a 700x speed-up.

We also directly compare all generated labels with human labels on the KITTI-360 training set (see~\cref{tab:pseudo-labels}). We show significantly improved accuracy of our pseudo-labels compared to previous auto-labelling methods~\cite{liu2024vsrd, zakharov2020autolabeling}, both in 3D and BEV.

\paragraph{Waymo.} In the second experiment, we evaluate our method on Waymo dataset~\cite{Sun_2020_CVPR}, where we again train MonoDETR~\cite{zhang2023monodetr} detector using only our pseudo-labels generated for Waymo training set and evaluate on Waymo validation set (see \cref{tab:waymo}). Despite using no human annotations, our method achieves competitive performance to methods using human annotations. Note that we are unable to directly compare to VSRD~\cite{liu2024vsrd} on this dataset, because the training set has 150k frames, and it is therefore practically infeasible to auto-label the data using VSRD (15 minutes per frame) in any reasonable time frame. 

\paragraph{Pseudo-labels for pre-training.}

In the following experiments, the labels generated by our auto-labelling pipeline are used as a pre-training step and then only a small fraction of human labels is used to further fine-tune the model.

\begin{table}
\centering
\footnotesize
\setlength{\tabcolsep}{2pt}
\begin{tabular}{l c | c c}
\hline
\textbf{Training}& \textbf{Fine-tuning} & \textbf{AP\textsubscript{BEV}} & \textbf{AP\textsubscript{3D}@0.5}\\
\textit{\color{darkgreen} pseudo labels} & \textit{\color{orange} human labels} & & \\
\hline
Ours & 1\% & 21.59 & 18.43 \\
- & 1\% & 15.55 & 13.81 \\ \hline
Ours & 5\% & 22.56 & 20.68 \\
- & 5\% & 21.08 & 18.12 \\ \hline
Ours & 10\% & 22.80 & 20.65 \\
- & 10\% & 21.69 & 18.53 \\ \hline
Ours & 15\% & \underline{23.20} & \underline{20.88} \\
- & 15\% & 22.63 & 20.60 \\ \hline
- & 100\% & \textbf{23.63} & \textbf{21.41} \\
\hline
\end{tabular}
\caption{Monocular 3D object detection (vehicle - Level 2) on Waymo validation set. Training on Waymo training set using pseudo-labels and fine-tuning on a fraction of human labels.}
\label{tab:waymo-fine-tuning}
\end{table}

First, in \cref{tab:fine-tuning} we show that pre-training on a larger dataset (KITTI-360) auto-labelled by our method and then fine-tuning on the target domain (KITTI) significantly improves the average precision when compared to simply training on the target domain from scratch. \textit{With only 25\% of human labels}, our method \textbf{outperforms} traditional fully-supervised detector trained from scratch with 100\% of human labels, and the gap grows as more human labels are used. Also note that our method significantly outperforms current state-of-the-art VSRD~\cite{liu2024vsrd} by a large margin.

Second, we show on a large-scale dataset (Waymo) that pre-training with our method can save human annotations costs. In this scenario, we assume the annotation budget is limited and therefore it's not possible to label all images in the training set. We show, that when only a fraction of images are labelled by a human -- e.g. only 5\%, using our pseudo-labels as a pre-training step leads to consistent improvements in detection accuracy at no annotation cost, compared to only using available human labels (see \cref{tab:waymo-fine-tuning}). Also, when using only 15\% of human labels and our method as a pre-training step, the accuracy is almost identical to using all human labels; this can also be viewed that using our method leads to \textit{saving 85\% of annotation costs} for a very small drop in accuracy.

\begin{table*}
\centering
\footnotesize
\setlength{\tabcolsep}{5pt}
\begin{tabular}{ l l | c c c | c c c }
\hline
\textbf{Dataset} & \textbf{Labels} & \multicolumn{3}{c|}{\textbf{AP\textsubscript{BEV}/AP\textsubscript{3D}@0.5}} & \multicolumn{3}{c}{\textbf{AP\textsubscript{BEV}/AP\textsubscript{3D}@0.3}} \\
\textbf{for training} & & Easy & Moderate & Hard & Easy & Moderate & Hard \\
\hline
KITTI & \color{darkgreen} pseudo & 61.24/53.22 & 45.66/40.88 & 38.67/34.68 & 77.69/76.07 & 63.39/58.20 & 55.70/54.46\\
K360 & \color{darkgreen} pseudo & 57.62/47.39 & 43.86/38.33 & 37.05/32.23 & 75.83/74.40 & 63.69/61.56 & 55.75/53.85 \\
KITTI + K360  & \color{darkgreen}pseudo & \underline{64.29}/\underline{58.97} & \underline{50.78}/\underline{44.23} & \underline{44.56}/\underline{40.58} & \underline{78.27}/\underline{77.71} & \underline{65.86}/\underline{64.85} & \textbf{62.35}/\underline{56.67} \\
\hline
KITTI & \color{red}human & \textbf{67.44}/\textbf{65.09} & \textbf{53.46}/\textbf{47.39} & \textbf{46.80}/\textbf{44.58} & \textbf{80.30}/\textbf{79.72} & \textbf{67.16}/\textbf{65.87} & \underline{59.54}/\textbf{58.83}\\
\hline 
\end{tabular}
\caption{MonoDETR~\cite{zhang2023monodetr} AP on the \textit{KITTI~\cite{geiger2013vision} validation} set, depending on which datasets/pseudo-labels are used in training.}
\label{tab:joint_training}
\end{table*}
\paragraph{Cross-dataset evaluation.}
Thanks to using Canonical Object Space, it is possible to combine datasets with different camera setups. First, we demonstrate that unlike previous work~\cite{liu2024vsrd}, our method works out-of-the-box in previously unseen camera setup (see \cref{tab:joint_training} - row K360 and \cref{tab:fine-tuning} - row 0\%), where the model was trained using pseudo-labels on KITTI-360 but evaluated on in this case unseen KITTI dataset. 
We also combine both KITTI and KITTI-360 datasets training subsets, generate our pseudo-labels, train a single model and evaluate it on the KITTI validation subset (\cref{tab:joint_training} -- row KITTI+K360). Combining pseudo-labels from KITTI+K360 achieves comparable accuracy to using human labels of KITTI, and \textit{even surpasses the accuracy of training with human labels} in the Hard category in the 0.3 metric. 

\begin{figure}
\centering
\renewcommand{\arraystretch}{0.3} 
\begin{tabular}{c}
    \includegraphics[width=0.95\linewidth]{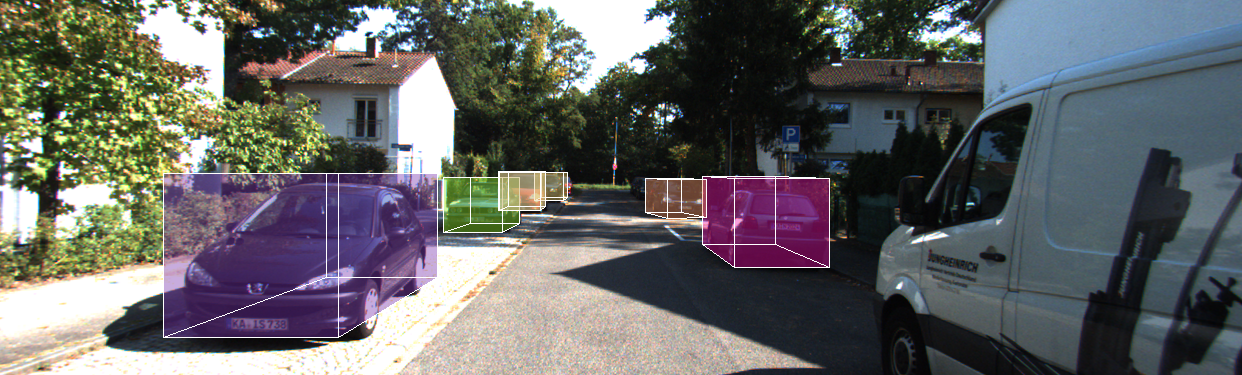} \\
    \includegraphics[width=0.95\linewidth]{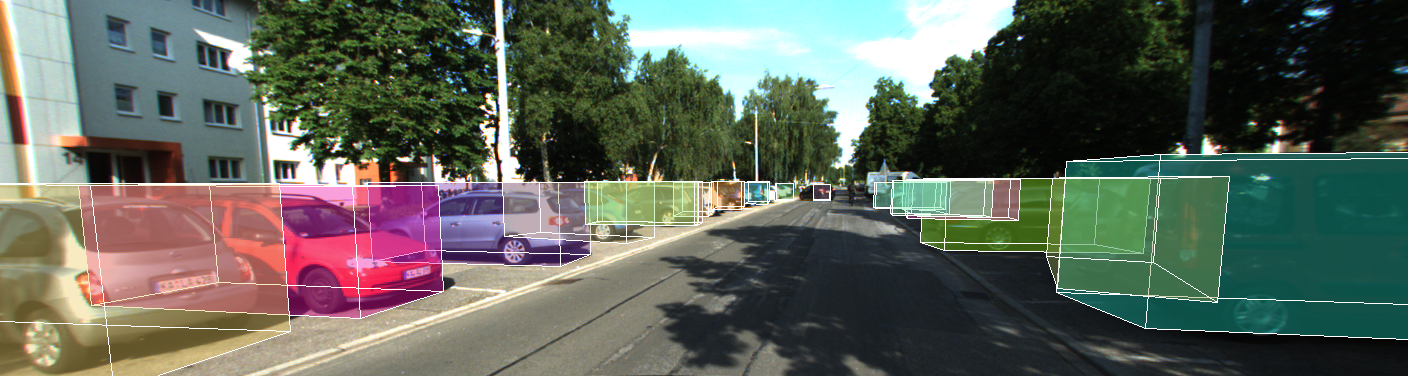} \\
    \includegraphics[width=0.95\linewidth, trim=0px 100px 0px 300px, clip]{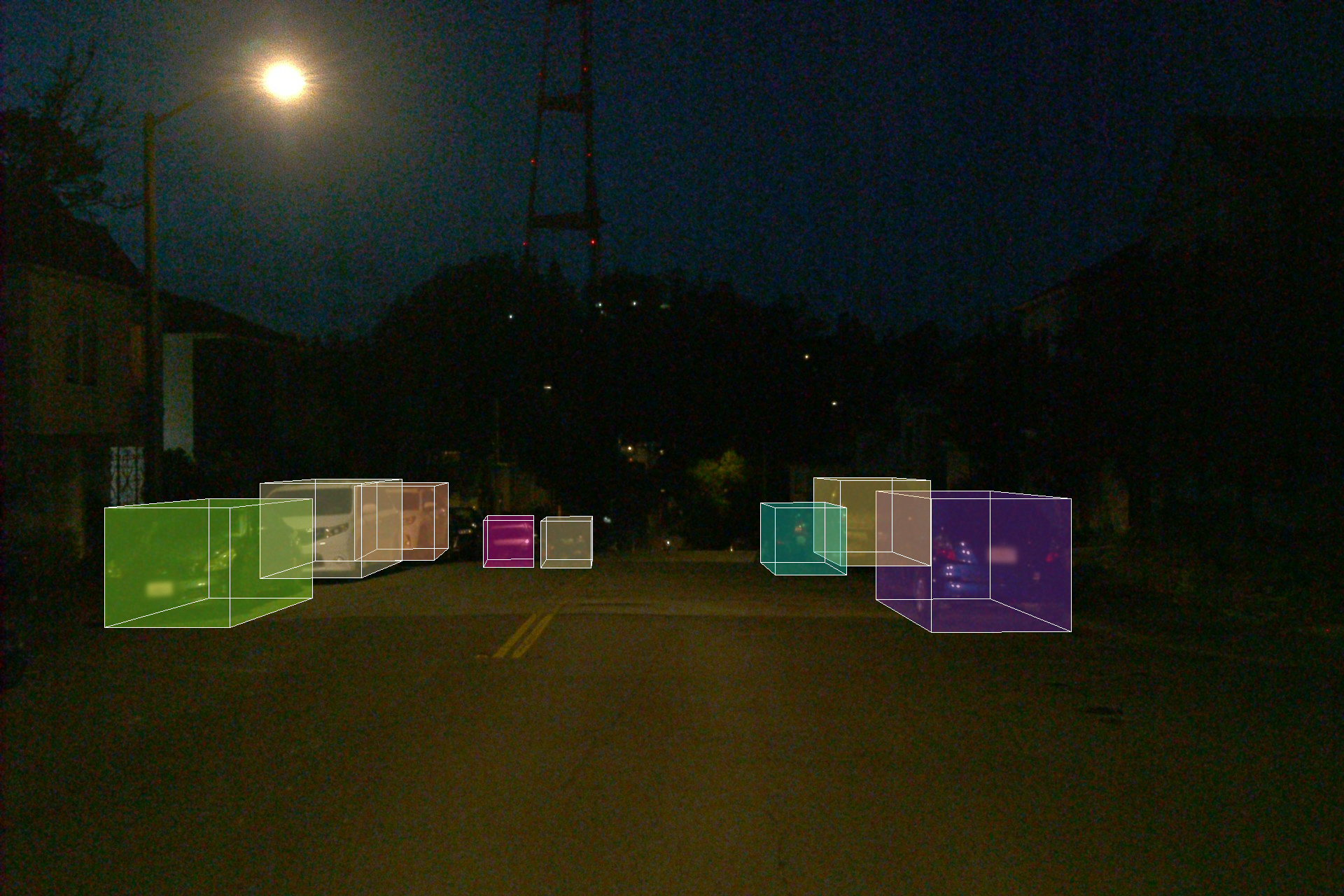}
\end{tabular}
\captionof{figure}{Example of 3D car detections by MonoDETR~\cite{zhang2023monodetr} trained without human annotations on KITTI~\cite{geiger2013vision} (top), KITTI-360~\cite{liao2022kitti} (middle) and Waymo~\cite{Sun_2020_CVPR} (bottom)}
\label{fig:qualitative_analysis}
\end{figure}
\paragraph{Qualitative analysis.}
We provide qualitative analysis in \cref{fig:qualitative_analysis} and in the Supplementary material in \cref{fig:qualitative_analysis_supp} and \cref{fig:qualitative_analysis_supp_waymo}.

\subsection{Ablations}
In this section, we present two ablation experiments we deemed the most important, and, due to lack of space, we kindly refer the reader to the Supplementary material for further ablations. The auto-labelling pipeline ablations use KITTI~\cite{geiger2013vision} training set for evaluation as it best measures the quality of the auto-generated labels, while the ablations involving a trained model (MonoDETR) use the respective KITTI training/validation sets for training and evaluation.

First, we inspect the impact of the proposed Local Object Motion Model and Saturated Closeness Criterion. As seen in ~\cref{tab:abl_criterion_pr}, both components are crucial to achieve the best accuracy of generated pseudo-labels. 
\begin{table}
\centering
\small
\begin{tabular}{ c c | c c c }
\hline
LOMM & SCC & \multicolumn{3}{c}{\textbf{AP\textsubscript{BEV}/AP\textsubscript{3D}@0.5}}\\
& & Easy & Moderate & Hard \\
\hline
\xmark & \xmark & 20.41/17.40 & 18.22/15.88 & 19.99/13.75 \\
\xmark & \checkmark & 20.37/17.16 & 18.26/15.76 & 20.04/13.66 \\
\checkmark & \xmark & \underline{35.89}/\underline{29.05} & \underline{26.32}/\underline{17.43} & \underline{28.87}/\underline{19.19} \\
\checkmark & \checkmark & \textbf{39.22}/\textbf{32.29} & \textbf{33.00}/\textbf{23.14} & \textbf{31.09}/\textbf{21.34} \\
\hline
\end{tabular}
\caption{Local Object Motion Model (LOMM) and Saturated Closeness Criterion (SCC) ablation on the KITTI~\cite{geiger2013vision} training set.}
\label{tab:abl_criterion_pr}
\end{table}
In the second ablation, we don't use Canonical Object Space and instead naively combine KITTI and KITTI-360 datasets. As expected, the average precision decreases when KITTI-360 data are appended to KITTI (see \cref{tab:abl_canonical}), because the network focuses on camera setup of the larger K360 dataset, and it is only with COS enabled that the network actually benefits from the added data. Note that only in this ablation, we use human labels to eliminate any effects of pseudo-labels.

\begin{table}
\centering
\small
\setlength{\tabcolsep}{4pt}
\begin{tabular}{lc|ccc}
\hline
\textbf{Dataset} &  & \multicolumn{3}{c}{\textbf{AP\textsubscript{BEV}/AP\textsubscript{3D}@0.3}} \\

\textbf{for training}& \textbf{COS} & Easy & Moderate & Hard \\
\hline
KITTI & \xmark & \underline{80.30}/\underline{79.72} & \textbf{67.16}/\underline{65.87} & \underline{59.54}/\underline{58.83} \\
KITTI+K360 & \xmark & 71.71/69.62 & 59.44/57.89 & 53.76/52.25 \\
KITTI+K360 & \checkmark & \textbf{82.07}/\textbf{81.25} & \underline{67.14}/\textbf{66.16}& \textbf{62.47}/\textbf{61.02} \\
\hline
\end{tabular}
\caption{Canonical Object Space (COS) ablation using human labels, evaluated on KITTI validation set.}
\label{tab:abl_canonical}
\end{table}

\section{Conclusion}

A novel method for training monocular 3D detectors without domain-specific human annotations was presented. The method exploits temporal consistency in video sequences to automatically create 3D labels of objects (cars) and, as such, does not require human labels or additional sensors such as LiDAR. The method uses newly proposed Local Object Motion Model (LOMM) to disentangle object movement source between subsequent frames, is approximately 700 times faster than current state of the art and is able to compensate for camera focal length differences to improve scalability across multiple datasets.

The method is evaluated on three large-scale public datasets, where despite using no human labels, it outperforms prior work by a significant margin both for BEV and 3D when using the stricter 0.5 IoU evaluation. The method also shows its versatility by being a powerful pre-training tool for fully-supervised monocular 3D detection, and last but not least, shows that combining pseudo-labels from multiple datasets can achieve comparable accuracy to using human labels from a single dataset. The main limitation is detecting objects far away, which is not primarily due to the auto-labelling procedure but due to the inherent limitation of the trained monocular detector to infer the distance of objects which are only a couple of pixels high.

\bibliographystyle{ieeetr}
\bibliography{main}

\begin{thebibliography}{10}

\bibitem{peng2021weakm3d}
L.~Peng, S.~Yan, B.~Wu, Z.~Yang, X.~He, and D.~Cai, ``Weakm3d: Towards weakly supervised monocular 3d object detection,'' in {\em International Conference on Learning Representations}, 2022.

\bibitem{zakharov2020autolabeling}
S.~Zakharov, W.~Kehl, A.~Bhargava, and A.~Gaidon, ``Autolabeling 3d objects with differentiable rendering of sdf shape priors,'' in {\em Proceedings of the IEEE/CVF Conference on Computer Vision and Pattern Recognition}, pp.~12224--12233, 2020.

\bibitem{tao2023weakly}
R.~Tao, W.~Han, Z.~Qiu, C.-Z. Xu, and J.~Shen, ``Weakly supervised monocular 3d object detection using multi-view projection and direction consistency,'' in {\em Proceedings of the IEEE/CVF Conference on Computer Vision and Pattern Recognition}, pp.~17482--17492, 2023.

\bibitem{liu2024vsrd}
Z.~Liu, H.~Sakuma, and M.~Okutomi, ``Vsrd: Instance-aware volumetric silhouette rendering for weakly supervised 3d object detection,'' in {\em Proceedings of the IEEE/CVF Conference on Computer Vision and Pattern Recognition}, pp.~17354--17363, 2024.

\bibitem{yin2023metric3d}
W.~Yin, C.~Zhang, H.~Chen, Z.~Cai, G.~Yu, K.~Wang, X.~Chen, and C.~Shen, ``Metric3d: Towards zero-shot metric 3d prediction from a single image,'' in {\em Proceedings of the IEEE/CVF International Conference on Computer Vision}, pp.~9043--9053, 2023.

\bibitem{brazil2023omni3d}
G.~Brazil, A.~Kumar, J.~Straub, N.~Ravi, J.~Johnson, and G.~Gkioxari, ``Omni3d: A large benchmark and model for 3d object detection in the wild,'' in {\em Proceedings of the IEEE/CVF conference on computer vision and pattern recognition}, pp.~13154--13164, 2023.

\bibitem{ranftl2020towards}
R.~Ranftl, K.~Lasinger, D.~Hafner, K.~Schindler, and V.~Koltun, ``Towards robust monocular depth estimation: Mixing datasets for zero-shot cross-dataset transfer,'' {\em IEEE transactions on pattern analysis and machine intelligence}, vol.~44, no.~3, pp.~1623--1637, 2020.

\bibitem{bhat2023zoedepth}
S.~F. Bhat, R.~Birkl, D.~Wofk, P.~Wonka, and M.~M{\"u}ller, ``Zoedepth: Zero-shot transfer by combining relative and metric depth,'' {\em arXiv preprint arXiv:2302.12288}, 2023.

\bibitem{yang2024depth}
L.~Yang, B.~Kang, Z.~Huang, X.~Xu, J.~Feng, and H.~Zhao, ``Depth anything: Unleashing the power of large-scale unlabeled data,'' in {\em Proceedings of the IEEE/CVF Conference on Computer Vision and Pattern Recognition}, pp.~10371--10381, 2024.

\bibitem{geiger2013vision}
A.~Geiger, P.~Lenz, C.~Stiller, and R.~Urtasun, ``Vision meets robotics: The kitti dataset,'' {\em The International Journal of Robotics Research}, vol.~32, no.~11, pp.~1231--1237, 2013.

\bibitem{hu2024metric3d}
M.~Hu, W.~Yin, C.~Zhang, Z.~Cai, X.~Long, H.~Chen, K.~Wang, G.~Yu, C.~Shen, and S.~Shen, ``Metric3d v2: A versatile monocular geometric foundation model for zero-shot metric depth and surface normal estimation,'' {\em arXiv preprint arXiv:2404.15506}, 2024.

\bibitem{oquab2023dinov2}
M.~Oquab, T.~Darcet, T.~Moutakanni, H.~Vo, M.~Szafraniec, V.~Khalidov, P.~Fernandez, D.~Haziza, F.~Massa, A.~El-Nouby, {\em et~al.}, ``Dinov2: Learning robust visual features without supervision,'' {\em arXiv preprint arXiv:2304.07193}, 2023.

\bibitem{liu2022convnet}
Z.~Liu, H.~Mao, C.-Y. Wu, C.~Feichtenhofer, T.~Darrell, and S.~Xie, ``A convnet for the 2020s,'' in {\em Proceedings of the IEEE/CVF conference on computer vision and pattern recognition}, pp.~11976--11986, 2022.

\bibitem{ranftl2021vision}
R.~Ranftl, A.~Bochkovskiy, and V.~Koltun, ``Vision transformers for dense prediction,'' in {\em Proceedings of the IEEE/CVF international conference on computer vision}, pp.~12179--12188, 2021.

\bibitem{weng2019monocular}
X.~Weng and K.~Kitani, ``Monocular 3d object detection with pseudo-lidar point cloud,'' in {\em Proceedings of the IEEE/CVF International Conference on Computer Vision Workshops}, pp.~0--0, 2019.

\bibitem{wang2019pseudo}
Y.~Wang, W.-L. Chao, D.~Garg, B.~Hariharan, M.~Campbell, and K.~Q. Weinberger, ``Pseudo-lidar from visual depth estimation: Bridging the gap in 3d object detection for autonomous driving,'' in {\em Proceedings of the IEEE/CVF conference on computer vision and pattern recognition}, pp.~8445--8453, 2019.

\bibitem{ma2019accurate}
X.~Ma, Z.~Wang, H.~Li, P.~Zhang, W.~Ouyang, and X.~Fan, ``Accurate monocular 3d object detection via color-embedded 3d reconstruction for autonomous driving,'' in {\em Proceedings of the IEEE/CVF international conference on computer vision}, pp.~6851--6860, 2019.

\bibitem{liu2020smoke}
Z.~Liu, Z.~Wu, and R.~T{\'o}th, ``Smoke: Single-stage monocular 3d object detection via keypoint estimation,'' in {\em Proceedings of the IEEE/CVF conference on computer vision and pattern recognition workshops}, pp.~996--997, 2020.

\bibitem{huang2022monodtr}
K.-C. Huang, T.-H. Wu, H.-T. Su, and W.~H. Hsu, ``Monodtr: Monocular 3d object detection with depth-aware transformer,'' in {\em Proceedings of the IEEE/CVF conference on computer vision and pattern recognition}, pp.~4012--4021, 2022.

\bibitem{zhang2023monodetr}
R.~Zhang, H.~Qiu, T.~Wang, Z.~Guo, Z.~Cui, Y.~Qiao, H.~Li, and P.~Gao, ``Monodetr: Depth-guided transformer for monocular 3d object detection,'' in {\em Proceedings of the IEEE/CVF International Conference on Computer Vision}, pp.~9155--9166, 2023.

\bibitem{vaswani2017attention}
A.~Vaswani, ``Attention is all you need,'' {\em Advances in Neural Information Processing Systems}, 2017.

\bibitem{he2023ssd}
X.~He, F.~Yang, K.~Yang, J.~Lin, H.~Fu, M.~Wang, J.~Yuan, and Z.~Li, ``Ssd-monodetr: Supervised scale-aware deformable transformer for monocular 3d object detection,'' {\em IEEE Transactions on Intelligent Vehicles}, 2023.

\bibitem{zhou2023monoatt}
Y.~Zhou, H.~Zhu, Q.~Liu, S.~Chang, and M.~Guo, ``Monoatt: Online monocular 3d object detection with adaptive token transformer,'' in {\em Proceedings of the IEEE/CVF Conference on Computer Vision and Pattern Recognition}, pp.~17493--17503, 2023.

\bibitem{zhang2021objects}
Y.~Zhang, J.~Lu, and J.~Zhou, ``Objects are different: Flexible monocular 3d object detection,'' in {\em Proceedings of the IEEE/CVF Conference on Computer Vision and Pattern Recognition}, pp.~3289--3298, 2021.

\bibitem{yan2024monocd}
L.~Yan, P.~Yan, S.~Xiong, X.~Xiang, and Y.~Tan, ``Monocd: Monocular 3d object detection with complementary depths,'' in {\em Proceedings of the IEEE/CVF Conference on Computer Vision and Pattern Recognition}, pp.~10248--10257, 2024.

\bibitem{qin2020weakly}
Z.~Qin, J.~Wang, and Y.~Lu, ``Weakly supervised 3d object detection from point clouds,'' in {\em Proceedings of the 28th ACM International Conference on Multimedia}, pp.~4144--4152, 2020.

\bibitem{meng2020ws3d}
Q.~Meng, W.~Wang, T.~Zhou, J.~Shen, L.~Van~Gool, and D.~Dai, ``Weakly supervised 3d object detection from lidar point cloud,'' in {\em ECCV}, 2020.

\bibitem{meng2021towards}
Q.~Meng, W.~Wang, T.~Zhou, J.~Shen, Y.~Jia, and L.~Van~Gool, ``Towards a weakly supervised framework for 3d point cloud object detection and annotation,'' {\em IEEE Transactions on Pattern Analysis and Machine Intelligence}, vol.~44, no.~8, pp.~4454--4468, 2021.

\bibitem{mccraith2022}
R.~McCraith, E.~Insafutdinov, L.~Neumann, and A.~Vedaldi, ``Lifting {2D} object locations to {3D} by discounting {LiDAR} outliers across objects and views,'' in {\em 2022 International Conference on Robotics and Automation (ICRA)}, pp.~2411--2418, 2022.

\bibitem{skvrna2025tcc}
J.~Skvrna and L.~Neumann, ``{TCC-Det}: Temporarily consistent cues for weakly-supervised {3D} detection,'' in {\em European Conference on Computer Vision}, pp.~129--145, Springer, 2024.

\bibitem{huang2024vgw3d}
K.-C. Huang, Y.-H. Tsai, and M.-H. Yang, ``Weakly supervised 3d object detection via multi-level visual guidance,'' in {\em ECCV}, 2024.

\bibitem{cen2021open}
J.~Cen, P.~Yun, J.~Cai, M.~Y. Wang, and M.~Liu, ``Open-set 3d object detection,'' in {\em 2021 International conference on 3D vision (3DV)}, pp.~869--878, IEEE, 2021.

\bibitem{wang20224d}
Y.~Wang, Y.~Chen, and Z.-X. ZHANG, ``4d unsupervised object discovery,'' {\em Advances in Neural Information Processing Systems}, vol.~35, pp.~35563--35575, 2022.

\bibitem{najibi2022motion}
M.~Najibi, J.~Ji, Y.~Zhou, C.~R. Qi, X.~Yan, S.~Ettinger, and D.~Anguelov, ``Motion inspired unsupervised perception and prediction in autonomous driving,'' in {\em European Conference on Computer Vision}, pp.~424--443, Springer, 2022.

\bibitem{you2022learning}
Y.~You, K.~Luo, C.~P. Phoo, W.-L. Chao, W.~Sun, B.~Hariharan, M.~Campbell, and K.~Q. Weinberger, ``Learning to detect mobile objects from lidar scans without labels,'' in {\em Proceedings of the IEEE/CVF Conference on Computer Vision and Pattern Recognition}, pp.~1130--1140, 2022.

\bibitem{luo2023reward}
K.~Luo, Z.~Liu, X.~Chen, Y.~You, S.~Benaim, C.~P. Phoo, M.~Campbell, W.~Sun, B.~Hariharan, and K.~Q. Weinberger, ``Reward finetuning for faster and more accurate unsupervised object discovery,'' {\em Advances in Neural Information Processing Systems}, vol.~36, pp.~13250--13266, 2023.

\bibitem{zhang2023towards}
L.~Zhang, A.~J. Yang, Y.~Xiong, S.~Casas, B.~Yang, M.~Ren, and R.~Urtasun, ``Towards unsupervised object detection from lidar point clouds,'' in {\em Proceedings of the IEEE/CVF Conference on Computer Vision and Pattern Recognition}, pp.~9317--9328, 2023.

\bibitem{seidenschwarz2024semoli}
J.~Seidenschwarz, A.~Osep, F.~Ferroni, S.~Lucey, and L.~Leal-Taix{\'e}, ``Semoli: What moves together belongs together,'' in {\em Proceedings of the IEEE/CVF Conference on Computer Vision and Pattern Recognition}, pp.~14685--14694, 2024.

\bibitem{baur2024liso}
S.~A. Baur, F.~Moosmann, and A.~Geiger, ``Liso: Lidar-only self-supervised 3d object detection,'' in {\em European Conference on Computer Vision}, pp.~253--270, Springer, 2024.

\bibitem{li2022mvitv2}
Y.~Li, C.-Y. Wu, H.~Fan, K.~Mangalam, B.~Xiong, J.~Malik, and C.~Feichtenhofer, ``Mvitv2: Improved multiscale vision transformers for classification and detection,'' in {\em Proceedings of the IEEE/CVF conference on computer vision and pattern recognition}, pp.~4804--4814, 2022.

\bibitem{zhang2017efficient}
X.~Zhang, W.~Xu, C.~Dong, and J.~M. Dolan, ``Efficient l-shape fitting for vehicle detection using laser scanners,'' in {\em 2017 IEEE Intelligent Vehicles Symposium (IV)}, pp.~54--59, IEEE, 2017.

\bibitem{liao2022kitti}
Y.~Liao, J.~Xie, and A.~Geiger, ``Kitti-360: A novel dataset and benchmarks for urban scene understanding in 2d and 3d,'' {\em IEEE Transactions on Pattern Analysis and Machine Intelligence}, vol.~45, no.~3, pp.~3292--3310, 2022.

\bibitem{kumar2022deviant}
A.~Kumar, G.~Brazil, E.~Corona, A.~Parchami, and X.~Liu, ``Deviant: Depth equivariant network for monocular 3d object detection,'' in {\em European Conference on Computer Vision}, pp.~664--683, Springer, 2022.

\bibitem{Sun_2020_CVPR}
P.~Sun, H.~Kretzschmar, X.~Dotiwalla, A.~Chouard, V.~Patnaik, P.~Tsui, J.~Guo, Y.~Zhou, Y.~Chai, B.~Caine, V.~Vasudevan, W.~Han, J.~Ngiam, H.~Zhao, A.~Timofeev, S.~Ettinger, M.~Krivokon, A.~Gao, A.~Joshi, Y.~Zhang, J.~Shlens, Z.~Chen, and D.~Anguelov, ``Scalability in perception for autonomous driving: Waymo open dataset,'' in {\em Proceedings of the IEEE/CVF Conference on Computer Vision and Pattern Recognition (CVPR)}, June 2020.

\bibitem{wei2021fgr}
Y.~Wei, S.~Su, J.~Lu, and J.~Zhou, ``Fgr: Frustum-aware geometric reasoning for weakly supervised 3d vehicle detection,'' in {\em 2021 IEEE International Conference on Robotics and Automation (ICRA)}, pp.~4348--4354, IEEE, 2021.

\bibitem{wu2019detectron2}
Y.~Wu, A.~Kirillov, F.~Massa, W.-Y. Lo, and R.~Girshick, ``Detectron2.'' \url{https://github.com/facebookresearch/detectron2}, 2019.

\bibitem{lin2014microsoft}
T.-Y. Lin, M.~Maire, S.~Belongie, J.~Hays, P.~Perona, D.~Ramanan, P.~Doll{\'a}r, and C.~L. Zitnick, ``Microsoft {COCO}: Common objects in context,'' in {\em Computer Vision--ECCV 2014: 13th European Conference, Zurich, Switzerland, September 6-12, 2014, Proceedings, Part V 13}, pp.~740--755, Springer, 2014.

\bibitem{loshchilov2017decoupled}
I.~Loshchilov, ``Decoupled weight decay regularization,'' {\em arXiv preprint arXiv:1711.05101}, 2017.

\bibitem{rusinkiewicz2001icp}
S.~Rusinkiewicz and M.~Levoy, ``Efficient variants of the icp algorithm,'' in {\em Proceedings Third International Conference on 3-D Digital Imaging and Modeling}, pp.~145--152, 2001.

\end{thebibliography}

\maketitlesupplementary

\section{Ablations}


\subsection{Runtime analysis}
We provide a runtime analysis of auto-labelling on the KITTI-360~\cite{liao2022kitti} dataset in \cref{tab:runtime}, where our method takes approximately 1.3 seconds to autolabel each frame. We ran our experiments on a single node with 1x AMD EPYC 7534 CPU and 1x NVIDIA A100 GPU.

\begin{table}[h]
\centering
\footnotesize
\begin{tabular}{ l | c}
\hline
\textbf{Stage} & \textbf{Run time / frame [ms]}\\
\hline
Metric3DV2-G~\cite{hu2024metric3d} & $792 \pm 2.2$ \\
MViTv2-H~\cite{li2022mvitv2} & $318 \pm 2.0$ \\
Frames aggregation & $37 \pm 19.8$ \\
Optimization & $164 \pm 276$ \\
\hline
Total (ours) & $1\,311 \pm 291.6$ \\
VSRD ~\cite{liu2024vsrd} & $\approx 900\,000$ \\
\hline
\end{tabular}

\caption{Auto-labelling runtime on KITTI-360~\cite{liao2022kitti} dataset}
\label{tab:runtime}
\end{table}

\subsection{Human-annotated masks} In order to make a fair comparison, we additionally provide an experiment where instead of MViTV2~\cite{li2022mvitv2} segmentation masks, we use segmentation masks created by humans to have the exact same setting as VSRD~\cite{liu2024vsrd}. The experiment uses the KITTI-360~\cite{liao2022kitti} dataset, where an instance segmentation mask is provided for each object in each frame. This helps in two ways. First, we employ masks for point extraction, and as they are ground-truth, no false positives or negatives are present. Second, as the segmentation also comes with instance id, tracking vehicles is trivial.

As shown in \Cref{tab:abl_gt_masks}, using ground-truth masks significantly increases the accuracy, especially on the 0.3 IoU both on BEV and 3D. It also demonstrates that when using the same inputs as VSRD, our method outperforms VSRD when using 0.3 IoU evaluation, while maintaining its superior performance at 0.5 IoU, both on BEV and 3D, therefore achieving \textit{state-of-the-art results in both metrics}.

\begin{table}[!h]
\centering
\footnotesize
\setlength{\tabcolsep}{2pt}
\begin{tabular}{ l l | c c | c c }
\hline
\textbf{Method} & \textbf{Masks} & \multicolumn{2}{c|}{\textbf{AP\textsubscript{BEV}/AP\textsubscript{3D}@0.5}} & \multicolumn{2}{c}{\textbf{AP\textsubscript{BEV}/AP\textsubscript{3D}@0.3}} \\
& & Easy & Hard & Easy & Hard \\
\hline
VSRD~\cite{liu2024vsrd} & \color{red}yes &  29.07/21.77 & 22.83/16.46 &\underline{58.40}/\textbf{50.86} & \underline{50.61}/43.45 \\
\hline
\textit{Ours} & \color{darkgreen}no & \underline{38.41}/\underline{29.98} & \textbf{35.26}/\textbf{27.56} & 50.84/42.72 & 49.22/\textbf{46.59} \\
\textit{Ours} & \color{red}yes & \textbf{38.90}/\textbf{34.44} & \underline{33.49}/\underline{26.93} & \textbf{59.41}/\underline{49.03} & \textbf{54.11}/\underline{43.84} \\
\hline
\end{tabular}
\caption{Ablation study on the effect of using human-annotated masks instead of 2D detector for generating pseudo-labels to train MonoDETR~\cite{zhang2023monodetr} on KITTI-360~\cite{liao2022kitti} training set and evaluating on KITTI-360 test set}
\label{tab:abl_gt_masks}
\end{table}

\subsection{LiDAR and pseudo-LiDAR}
We further provide an analysis of how using LiDAR would affect our method's performance. Although monocular metric depth estimation has advanced significantly, it is still far from perfect. In \Cref{tab:abL_lidar}, we present a comparison of our method using either original LiDAR scans captured by Velodyne HDL-64E or using pseudo-LiDAR, which is created by lifting depth predictions from Metric3Dv2~\cite{hu2024metric3d} as in our method. Using original LiDAR scans significantly improves the average precision in both 0.3 and 0.5 IoU on Bird's Eye View (BEV) and 3D. LiDAR scans not only do not suffer from depth prediction errors but because of their consistency between frames, our method can use them to refine imprecise transformations between frames by employing the Iterative Closest Points (ICP) algorithm~\cite{rusinkiewicz2001icp}. Unfortunately, for pseudo-LiDAR, the ICP algorithm diverges in some cases.

To decrease the effect of the inaccurate pseudo-lidar, we speculate that using low-quality LiDAR generating sparse point clouds could guide and refine the pseudo-lidar, which is, on the other hand, very dense. Using a combination of LiDAR and pseudo-lidar might also mitigate the disadvantages of both methods.

\subsection{Number of frames used in aggregation}
\cref{tab:abl_frames} shows the ablation study on the number of frames used in the aggregation process. Perhaps surprisingly, aggregating over longer sequences does not yield better performance. 
 
We speculate that the distant car poses a significant challenge for the detector itself; therefore, creating precise 3D labels for such objects does not translate to better 3D detection accuracy.
\begin{table}[!h]
\centering
\small
\begin{tabular}{ c | c c c }
\hline
\textbf{No. of}& \multicolumn{3}{c}{\textbf{AP\textsubscript{BEV}/AP\textsubscript{3D}@0.5}}\\
\textbf{frames} & Easy & Moderate & Hard \\
\hline
$\pm$ 1 & 20.41/17.40 & 18.22/15.88 & 19.99/13.75 \\
$\pm$ 10 & 39.60/\textbf{34.37} & 32.85/\textbf{24.72} & 30.53/22.25 \\
$\pm$ 20 & \underline{39.77}/\underline{33.62} & \underline{33.89}/\textbf{24.72} & \underline{31.76}/\textbf{22.59} \\
$\pm$ 30 & \textbf{40.21}/33.05 & \textbf{34.24}/\underline{24.43} & \textbf{32.10}/\underline{22.33} \\
$\pm$ 50 & 39.27/32.37 & 33.33/23.44 & 31.35/21.57 \\
$\pm$ 100 & 39.23/32.19 & 28.29/23.05 & 31.14/21.25 \\
\hline
\end{tabular}

\caption{Number of frames used in the aggregation process ablation evaluating pseudo-labels directly on KITTI~\cite{geiger2013vision} training set.}
\label{tab:abl_frames}
\end{table}
\begin{table*}
\centering
\footnotesize
\setlength{\tabcolsep}{2pt}
\begin{tabular}{ l l | c c c | c c c }
\hline
\textbf{LiDAR} & \textbf{Labels} & \multicolumn{3}{c|}{\textbf{AP\textsubscript{BEV}/AP\textsubscript{3D}@0.5}} & \multicolumn{3}{c}{\textbf{AP\textsubscript{BEV}/AP\textsubscript{3D}@0.3}} \\
\textbf{for training} & & Easy & Moderate & Hard & Easy & Moderate & Hard \\
\hline
\color{darkgreen} no & \color{darkgreen} pseudo & 59.76/51.55 & 44.08/37.09 & 36.99/33.15 & 73.38/72.70 & 57.23/56.30 & 48.59/47.70\\
\color{red} yes & \color{darkgreen} pseudo & \underline{64.50}/\underline{60.04} & \underline{51.44}/\underline{44.50} & \underline{44.48}/\underline{37.76} & \underline{78.64}/\underline{77.62} & \underline{65.76}/\underline{63.79} & \underline{57.91}/\underline{56.36} \\
\hline
\color{red} yes & \color{red}human & \textbf{67.44}/\textbf{65.09} & \textbf{53.46}/\textbf{47.39} & \textbf{46.80}/\textbf{44.58} & \textbf{80.30}/\textbf{79.72} & \textbf{67.16}/\textbf{65.87} & \textbf{59.54}/\textbf{58.83}\\
\hline
\end{tabular}
\caption{Ablation study on the effect of using LiDAR scans to generate pseudo-labels to train MonoDETR~\cite{zhang2023monodetr} on KITTI~\cite{geiger2013vision} training set and evaluate on KITTI validation set.}
\label{tab:abL_lidar}
\end{table*}

\subsection{Steepness parameter in Saturated Closeness Criterion}
In this ablation, we explore possible values for the steepness parameter $\alpha$ in the Saturated Closeness Criterion, which controls what pseudo-LIDAR points are discarded as outliers. As shown in \cref{tab:abl_steepness}, $\alpha$ equal to 10 achieves the best performance.
\begin{table}[h]
\centering
\small
\begin{tabular}{ c | c c c }
\hline
& \multicolumn{3}{c}{\textbf{AP\textsubscript{BEV}/AP\textsubscript{3D}@0.5}}\\
$\mathbf{\alpha}$ & Easy & Moderate & Hard \\
\hline
1 & 39.19/32.07 & 28.28/23.05 & 31.13/21.24 \\
5 & \underline{39.33}/32.10 & \underline{28.31}/23.04 & \underline{31.15}/21.27 \\
10 & \textbf{39.38}/\textbf{32.26} & \textbf{33.10}/\textbf{23.16} & \textbf{31.20}/\textbf{21.36} \\
15 & 39.28/32.17 & 28.28/23.05 & 31.14/21.26 \\
20 & 39.23/\underline{32.25} & 28.20/\underline{23.11} & 31.10/\underline{21.33} \\
\hline
\end{tabular}

\caption{Ablation of how steepness parameter $\alpha$ affects AP of pseudo-labels on KITTI~\cite{geiger2013vision} training set.}
\label{tab:abl_steepness}
\end{table}

\begin{table}
\centering
\small
\begin{tabular}{ c | c c c }
\hline
\textbf{Canonical} & \multicolumn{3}{c}{\textbf{AP\textsubscript{BEV}/AP\textsubscript{3D}@0.5}}\\
\textbf{Focal Length} & Easy & Moderate & Hard \\
\hline
250 & \textbf{64.28}/\underline{55.49} & 49.10/41.98 & 42.35/35.64 \\
500 & 63.43/53.46 & 49.22/\underline{42.24} & 42.50/\underline{35.91} \\
750 & \underline{63.71}/\textbf{56.78} & \textbf{50.10}/\textbf{43.75}& \textbf{43.66}/\textbf{37.00} \\
1000 & 62.79/55.18 & \underline{49.39}/42.13 & \underline{42.80}/35.87 \\
\hline
\end{tabular}

\caption{Ablation study of how canonical focal length affects MonoDETR~\cite{zhang2023monodetr} AP trained on both KITTI~\cite{geiger2013vision} and KITTI-360~\cite{liao2022kitti} training set and evaluated on KITTI validation set}
\label{tab:abl_canonical_focal_length}
\end{table}
\subsection{Canonical Focal Length}
We also provide ablation on how the value of the canonical focal length affects the performance of MonoDETR~\cite{zhang2023monodetr} in \cref{tab:abl_canonical_focal_length}. The results show that the choice of focal length is not crucial as the performance does not change significantly between the chosen values. However, it is advisable to keep the canonical focal length close to the focal length of the cameras, as 500 and 750 achieve the best results and are also the closest to the real focal length of the cameras.

\section{Qualitative analysis}
\begin{figure*}
\centering
\renewcommand{\arraystretch}{0.3} 
\begin{tabular}{c c}
    \includegraphics[width=0.5\linewidth]{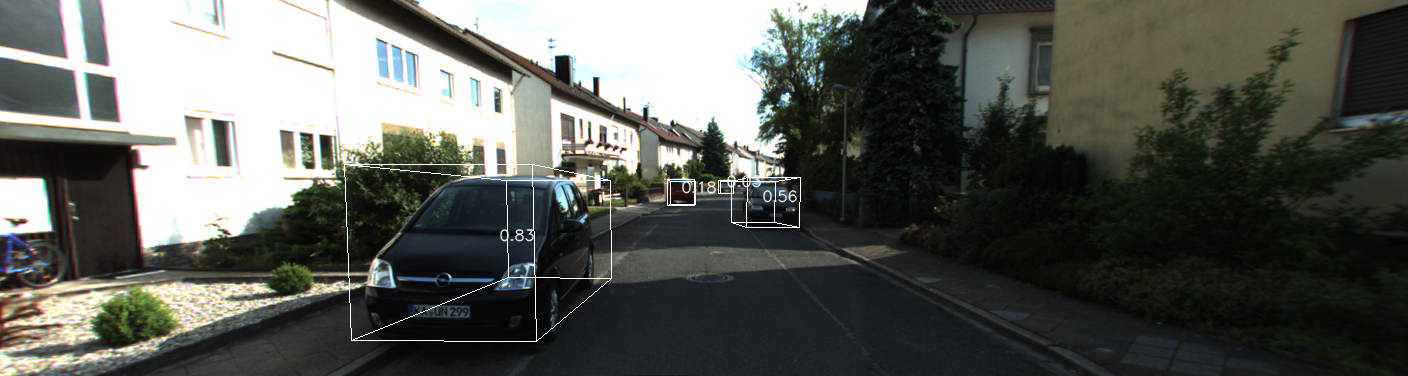} & \includegraphics[width=0.5\linewidth]{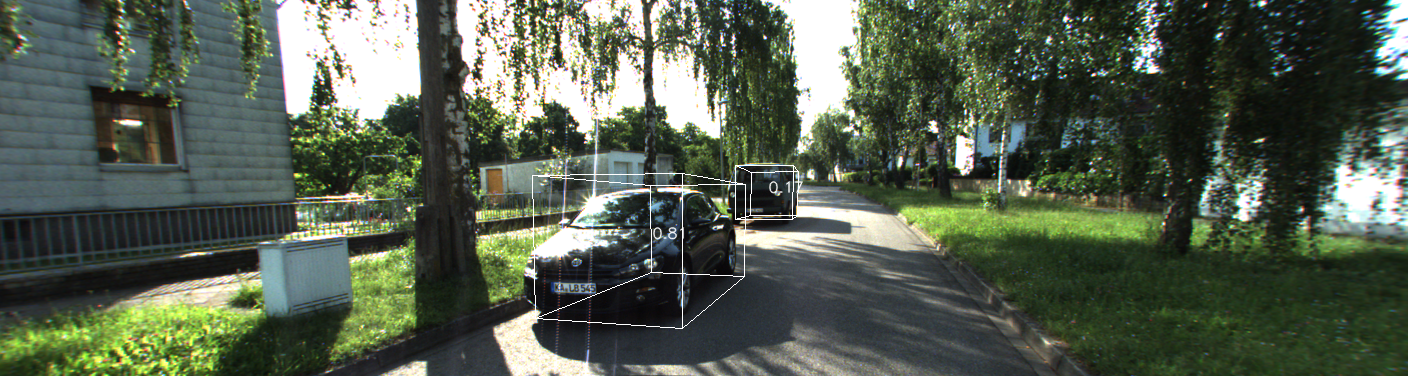} \\
    \includegraphics[width=0.5\linewidth]{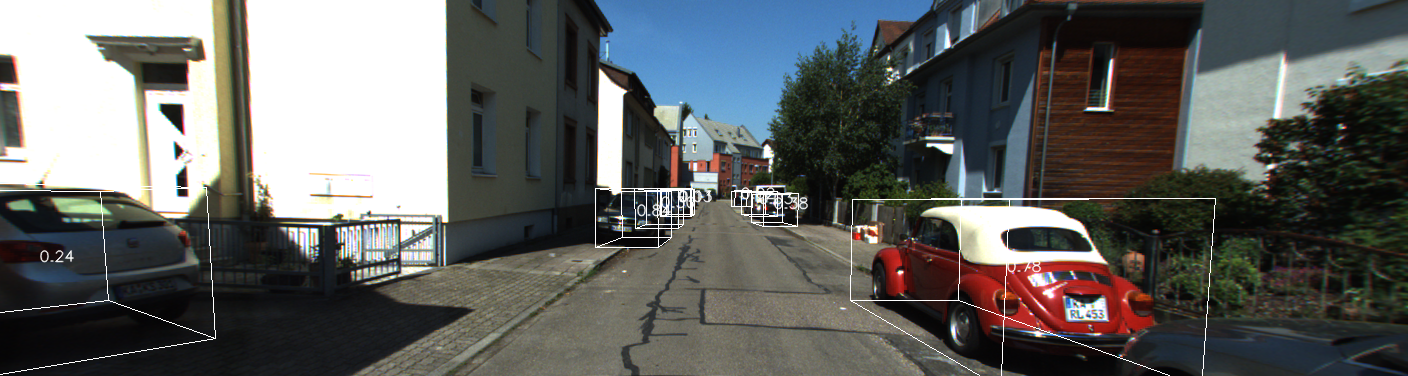} & \includegraphics[width=0.5\linewidth]{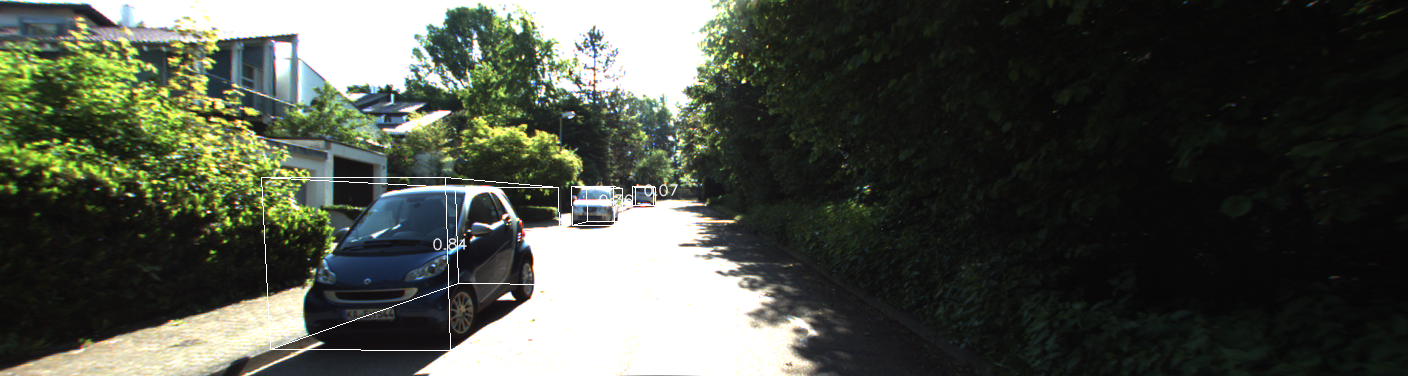} \\
    \includegraphics[width=0.5\linewidth]{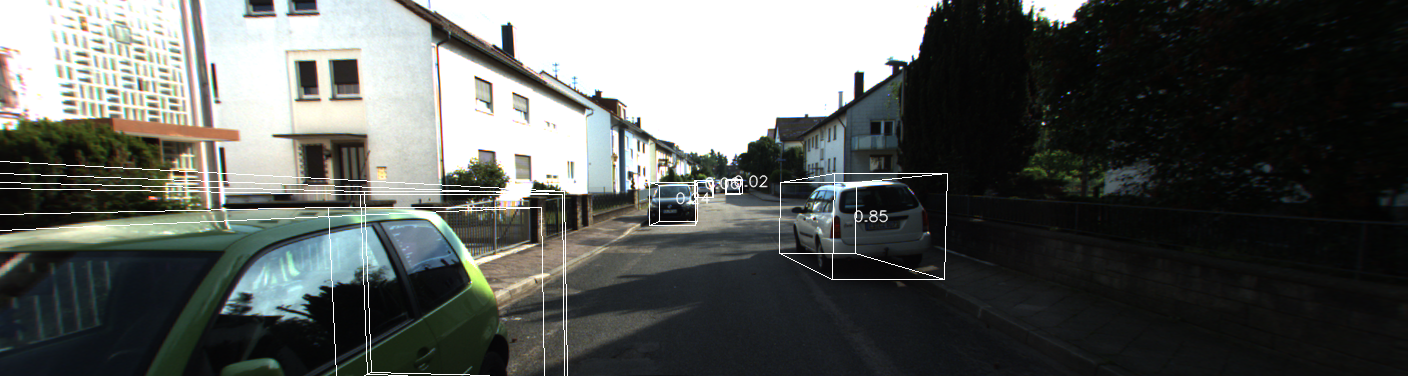} & \includegraphics[width=0.5\linewidth]{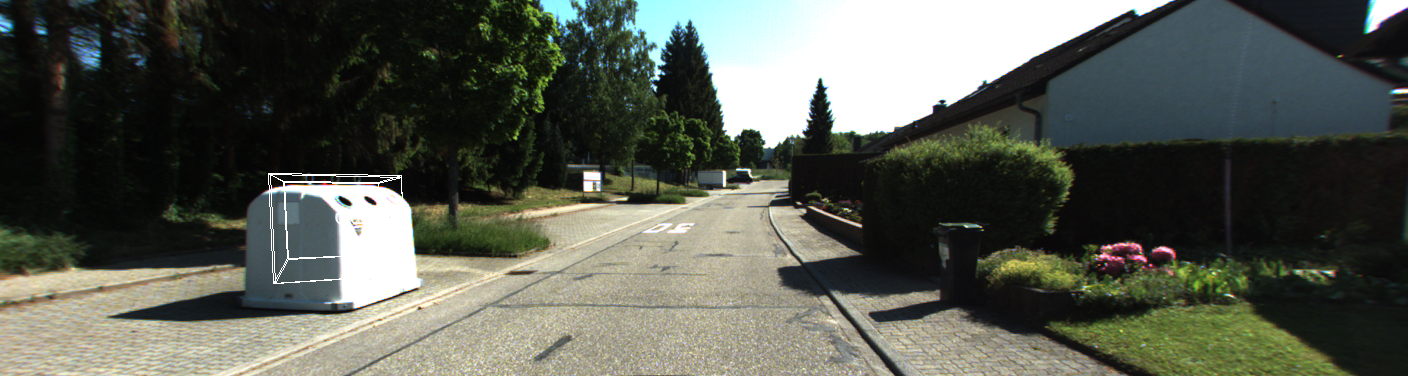} \\
    \includegraphics[width=0.5\linewidth]{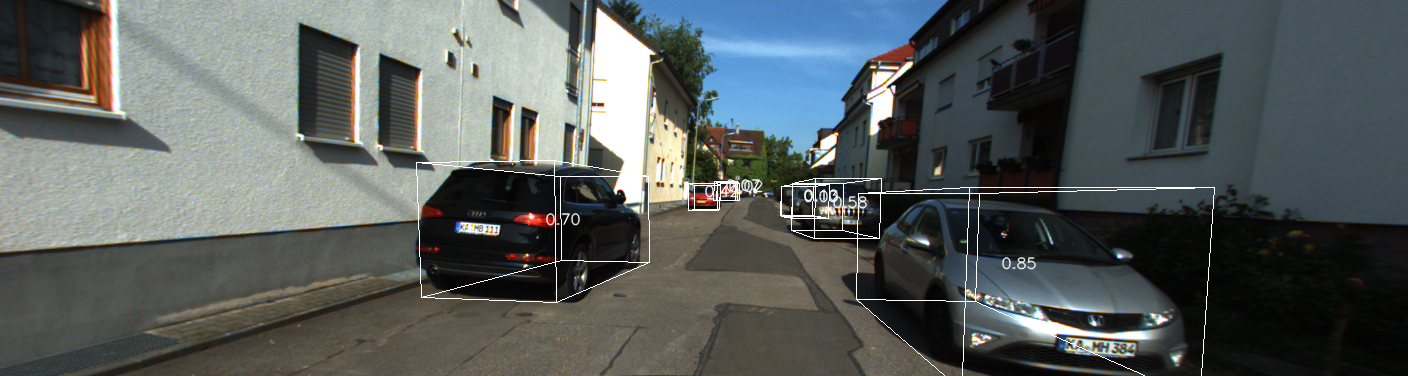} & \includegraphics[width=0.5\linewidth]{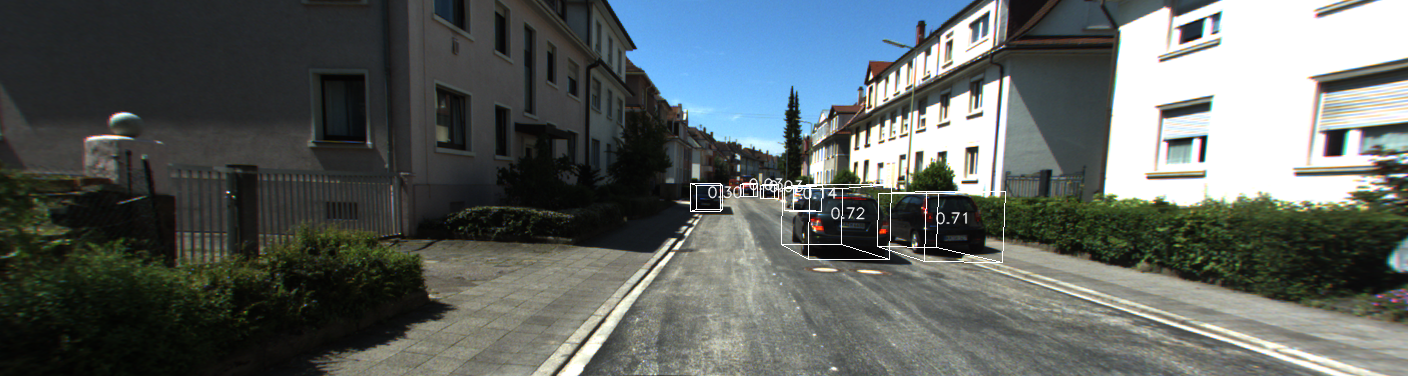} \\
    \includegraphics[width=0.5\linewidth]{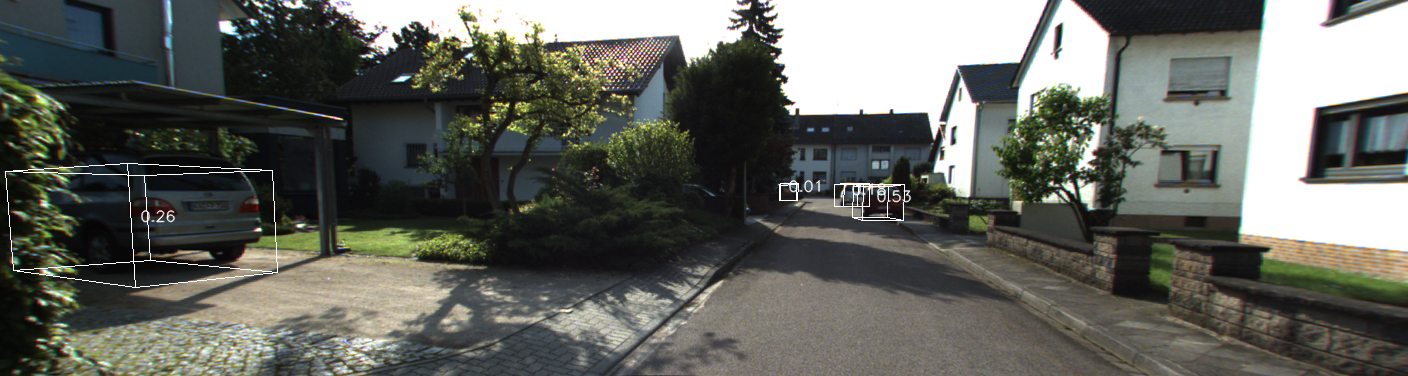} & \includegraphics[width=0.5\linewidth]{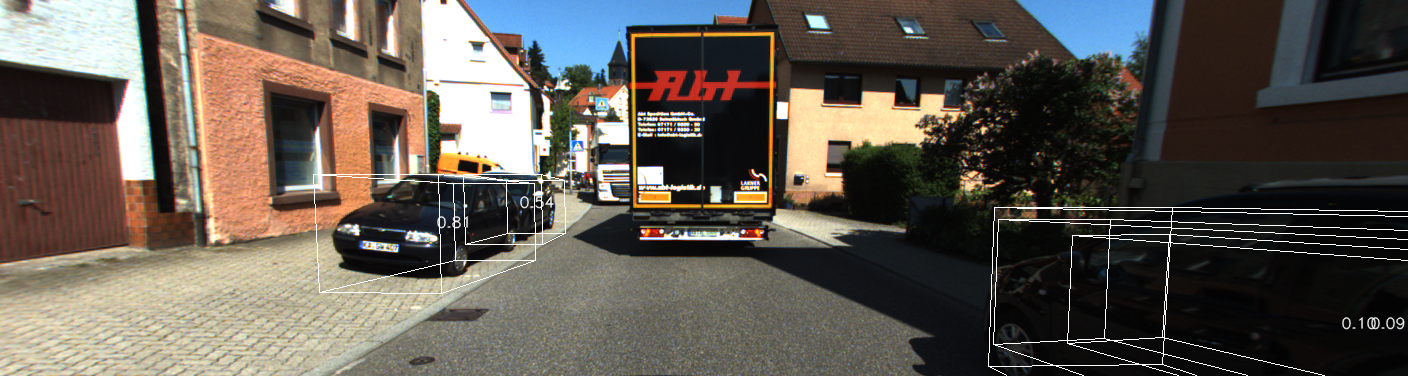} \\
    \includegraphics[width=0.5\linewidth]{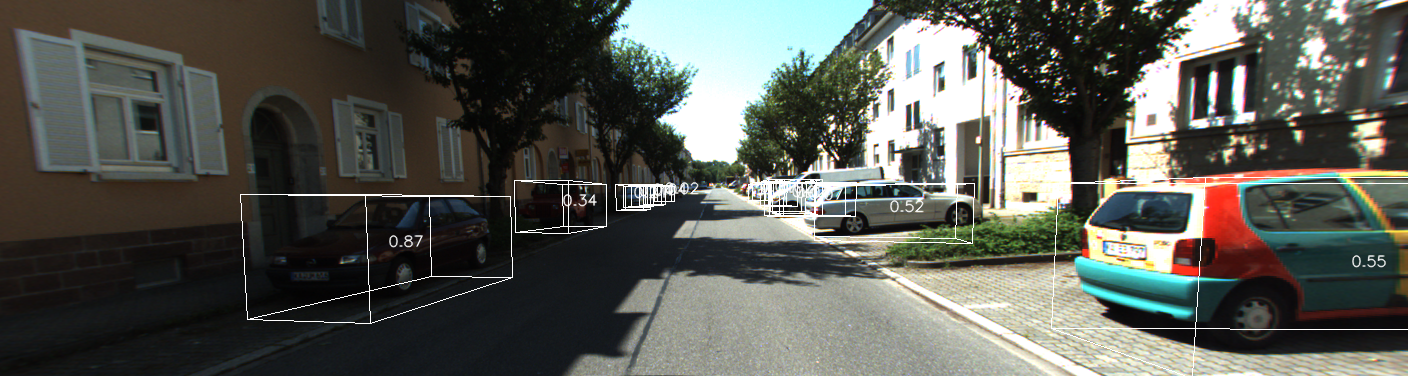} & \includegraphics[width=0.5\linewidth]{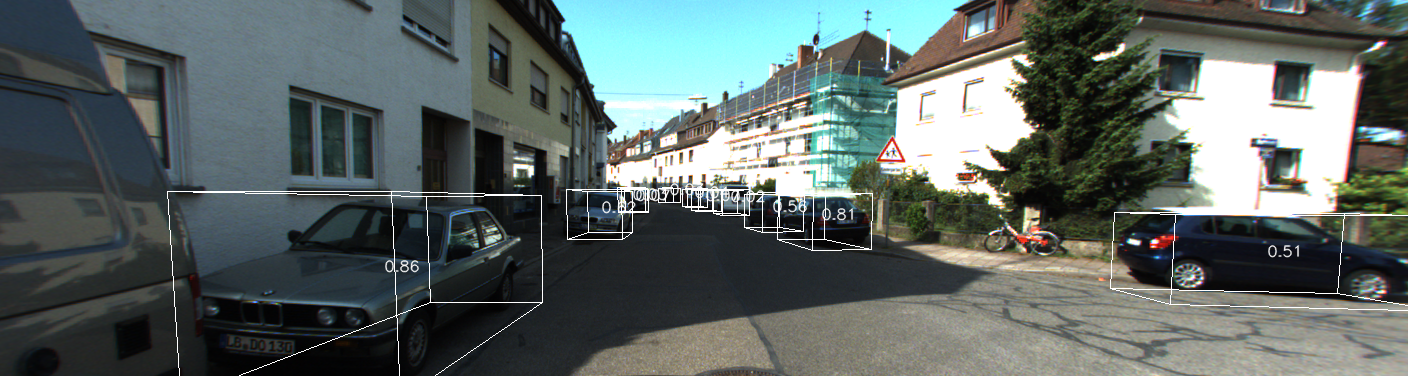} \\
    \includegraphics[width=0.5\linewidth]{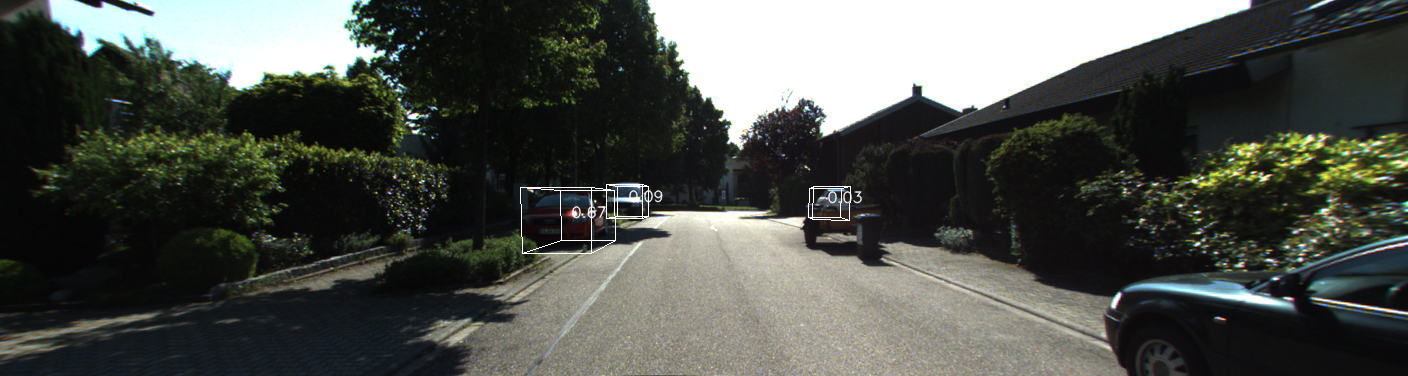} & \includegraphics[width=0.5\linewidth]{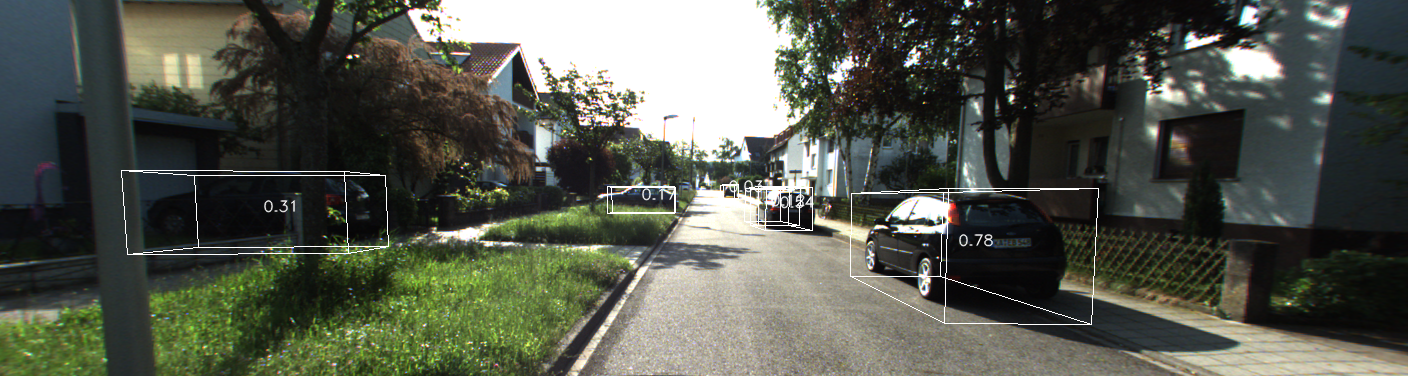} \\
    \includegraphics[width=0.5\linewidth]{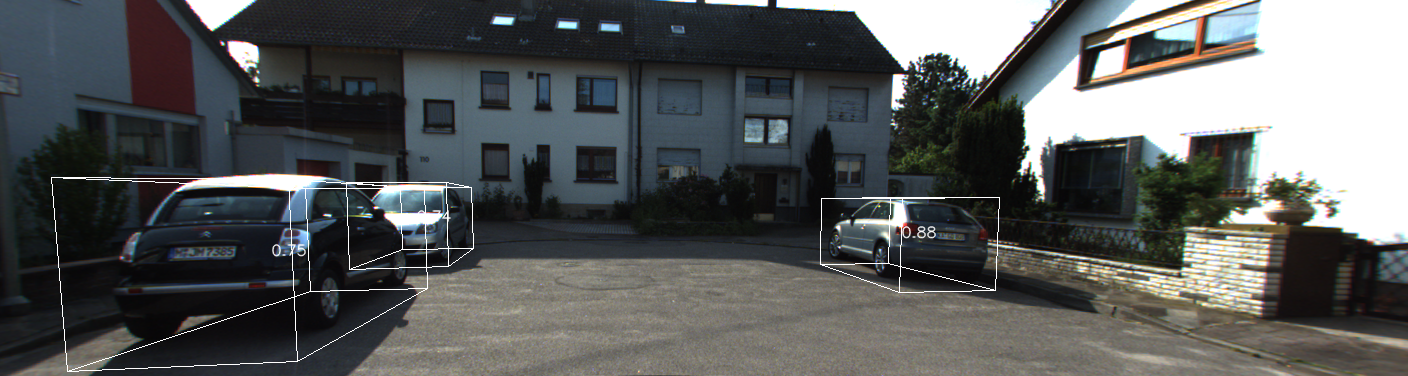} & \includegraphics[width=0.5\linewidth]{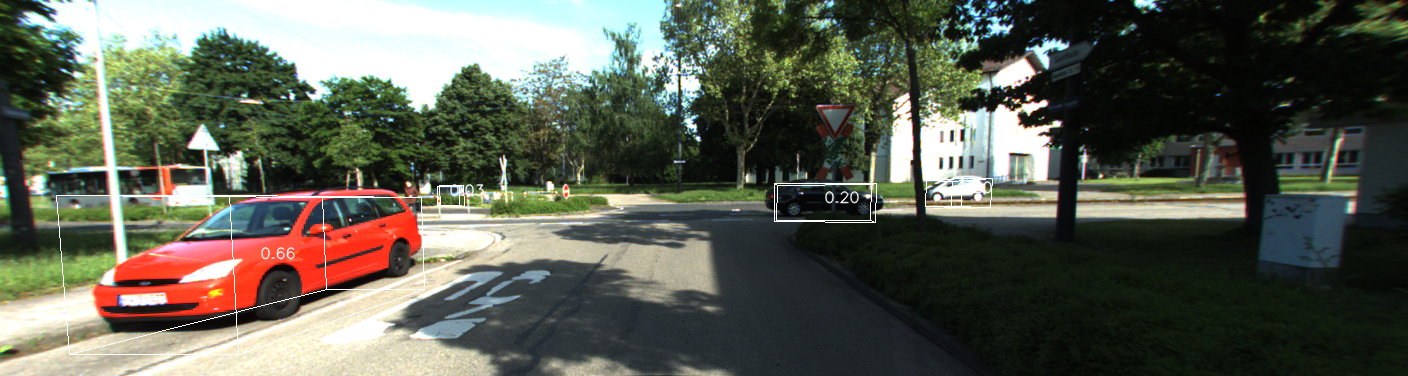} \\
    \includegraphics[width=0.5\linewidth]{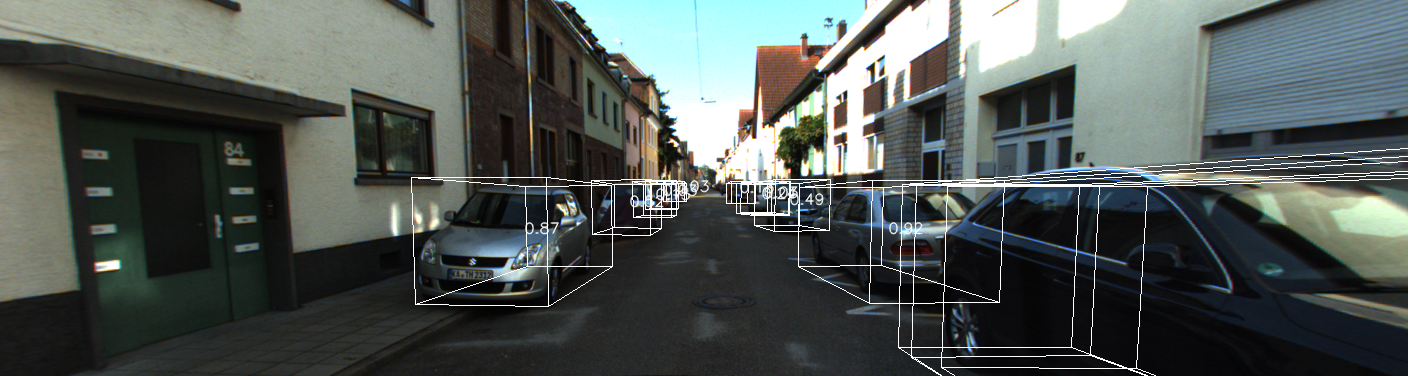} & \includegraphics[width=0.5\linewidth]{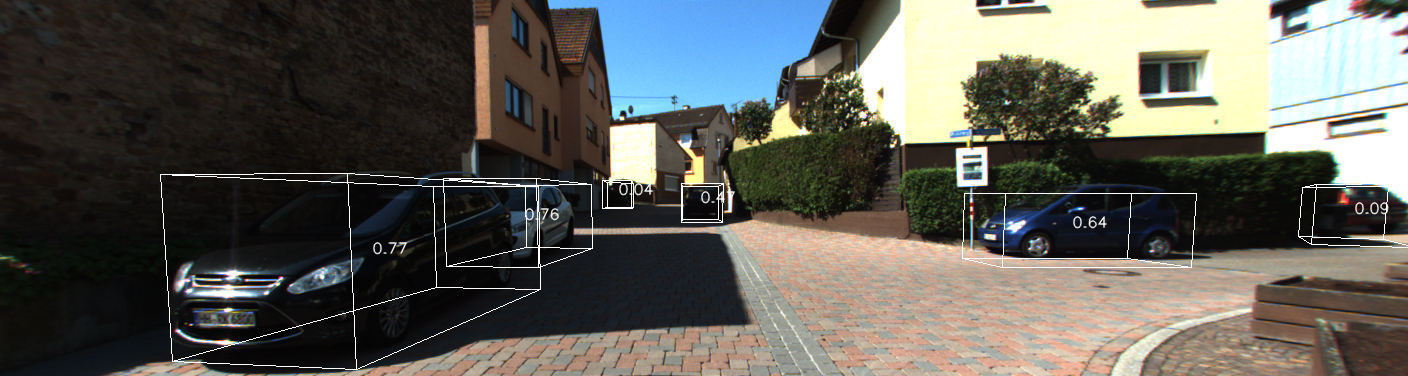} \\
\end{tabular}
\caption{Qualitative analysis of MonoDETR~\cite{zhang2023monodetr} trained using our method without using human annotations on KITTI~\cite{geiger2013vision} and KITTI-360~\cite{liao2022kitti} . White 3D bounding boxes are predictions, the number inside/near the bounding box is the confidence of each prediction.}
\label{fig:qualitative_analysis_supp}
\end{figure*}

In \Cref{fig:qualitative_analysis_supp}, we show multiple frames, in which we show how the MonoDETR~\cite{zhang2023monodetr} trained without human annotations performs on the KITTI-360~\cite{liao2022kitti} dataset. Please note that we are using our best model trained on both KITTI~\cite{geiger2013vision} and KITTI-360 datasets. We show both good and bad predictions, for example, the trash bin being recognized as a car. 

\begin{figure*}
\centering
\renewcommand{\arraystretch}{0.3} 
\begin{tabular}{c c}
    \includegraphics[width=0.5\linewidth, trim=0px 0px 0px 100px, clip]{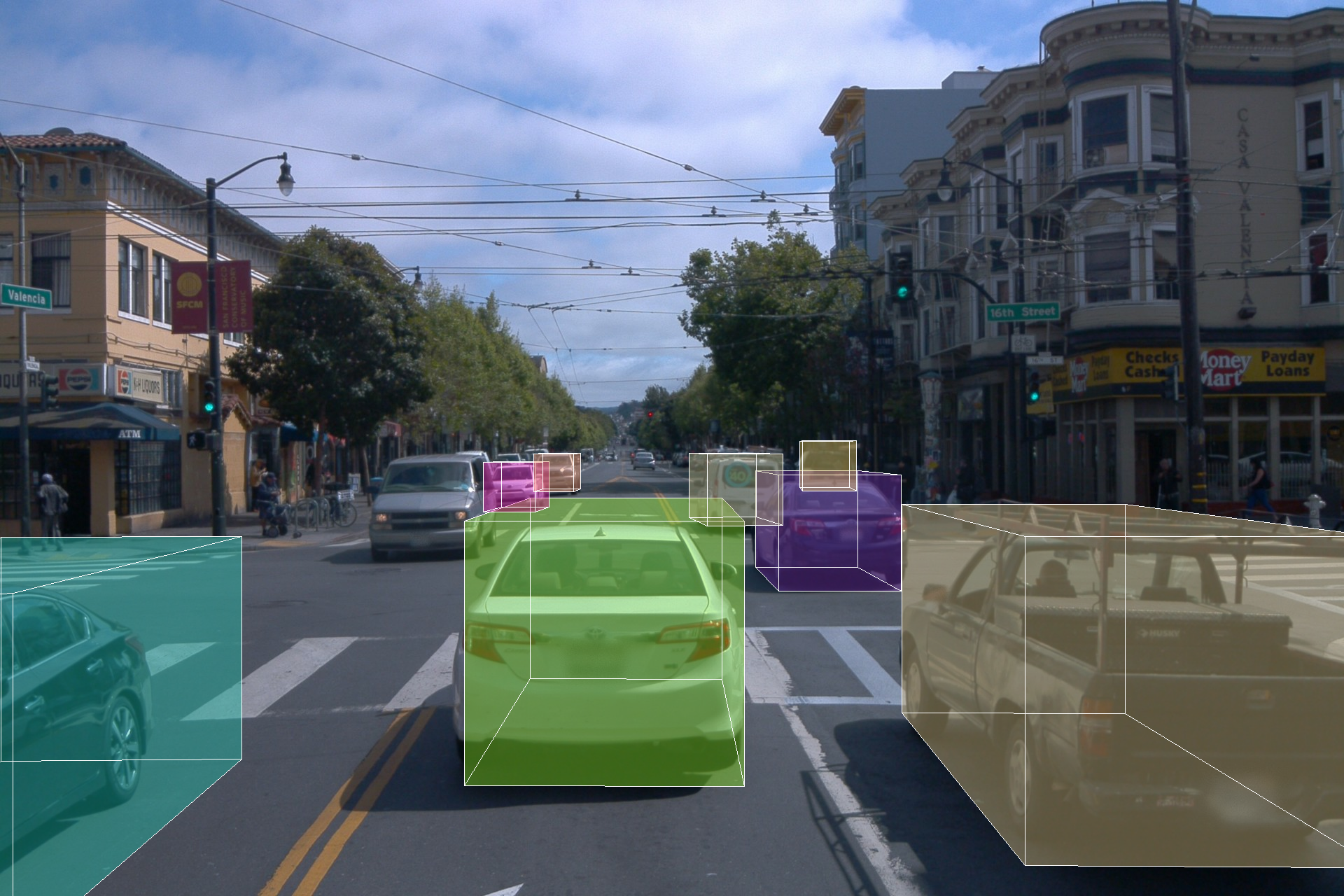} & \includegraphics[width=0.5\linewidth, trim=0px 0px 0px 100px, clip]{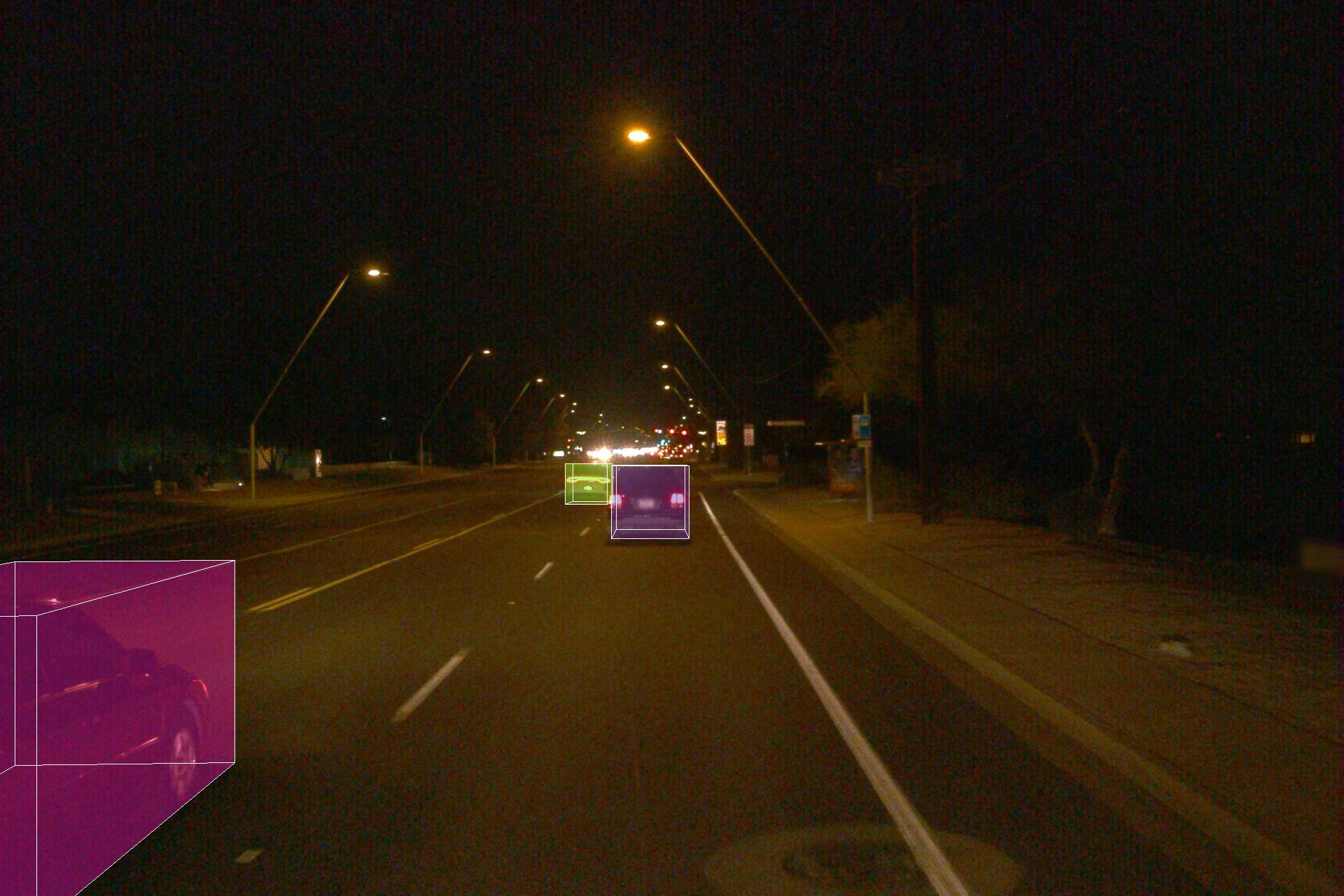} \\
    \includegraphics[width=0.5\linewidth, trim=0px 0px 0px 100px, clip]{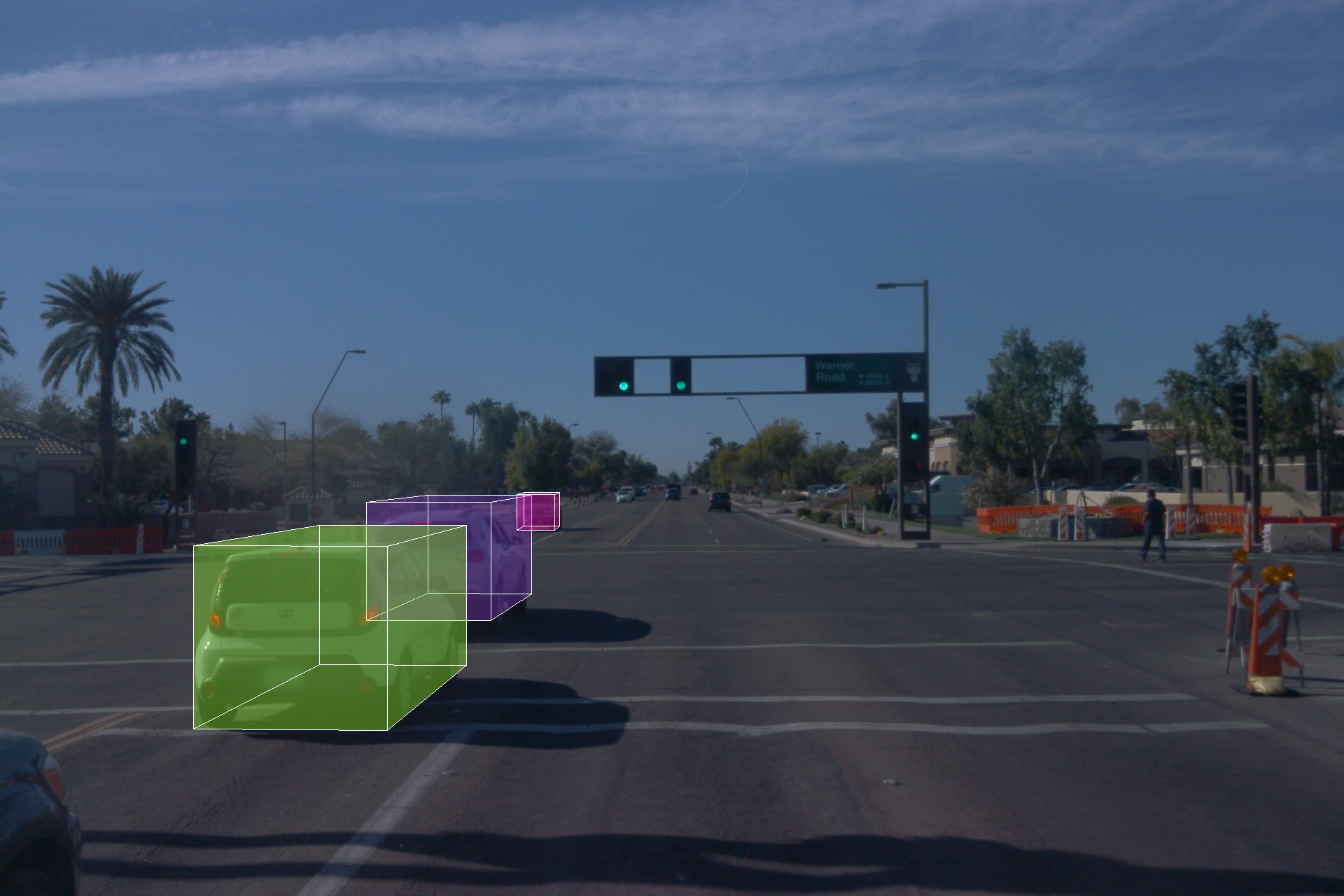} & \includegraphics[width=0.5\linewidth, trim=0px 0px 0px 100px, clip]{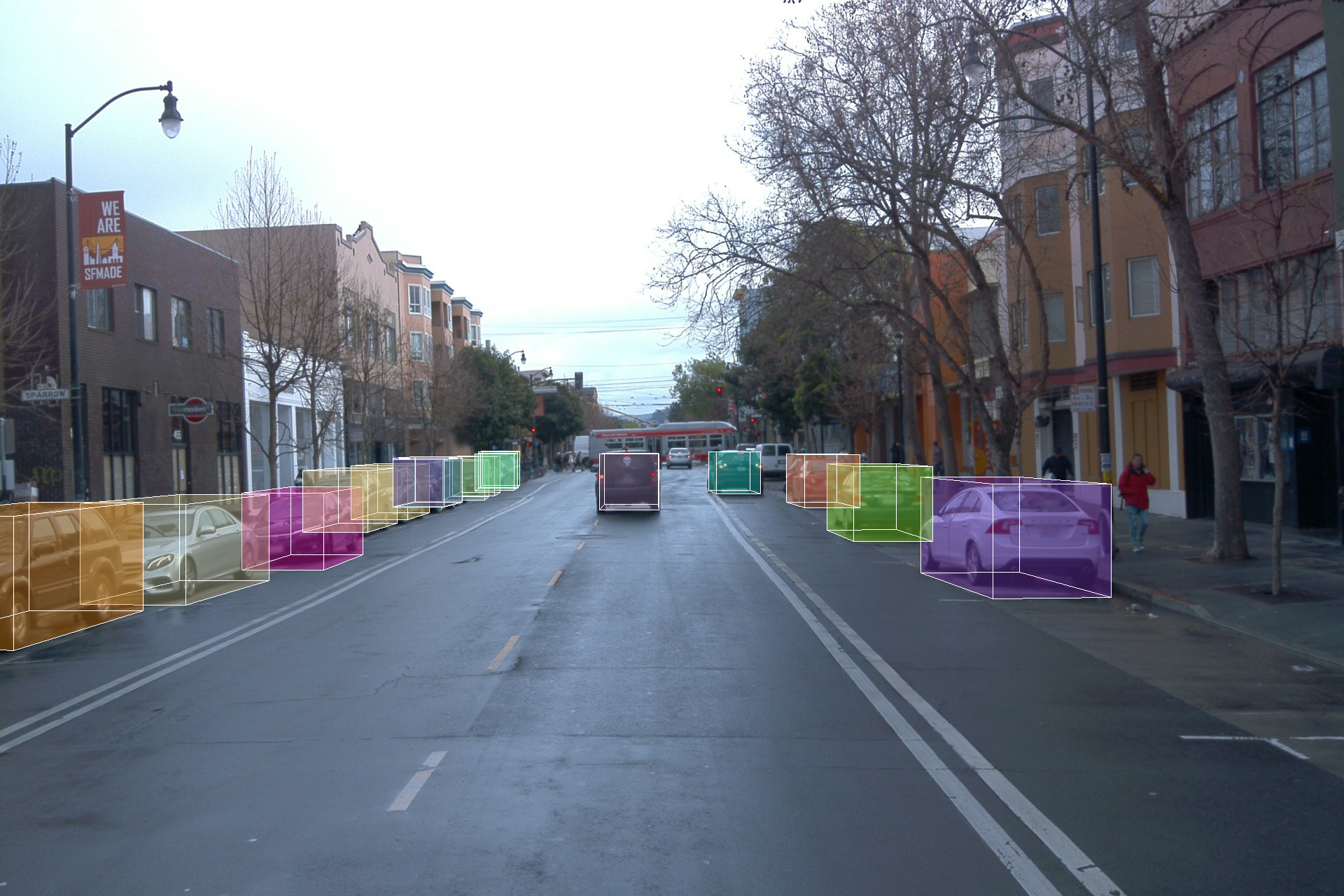} \\
    \includegraphics[width=0.5\linewidth, trim=0px 0px 0px 100px, clip]{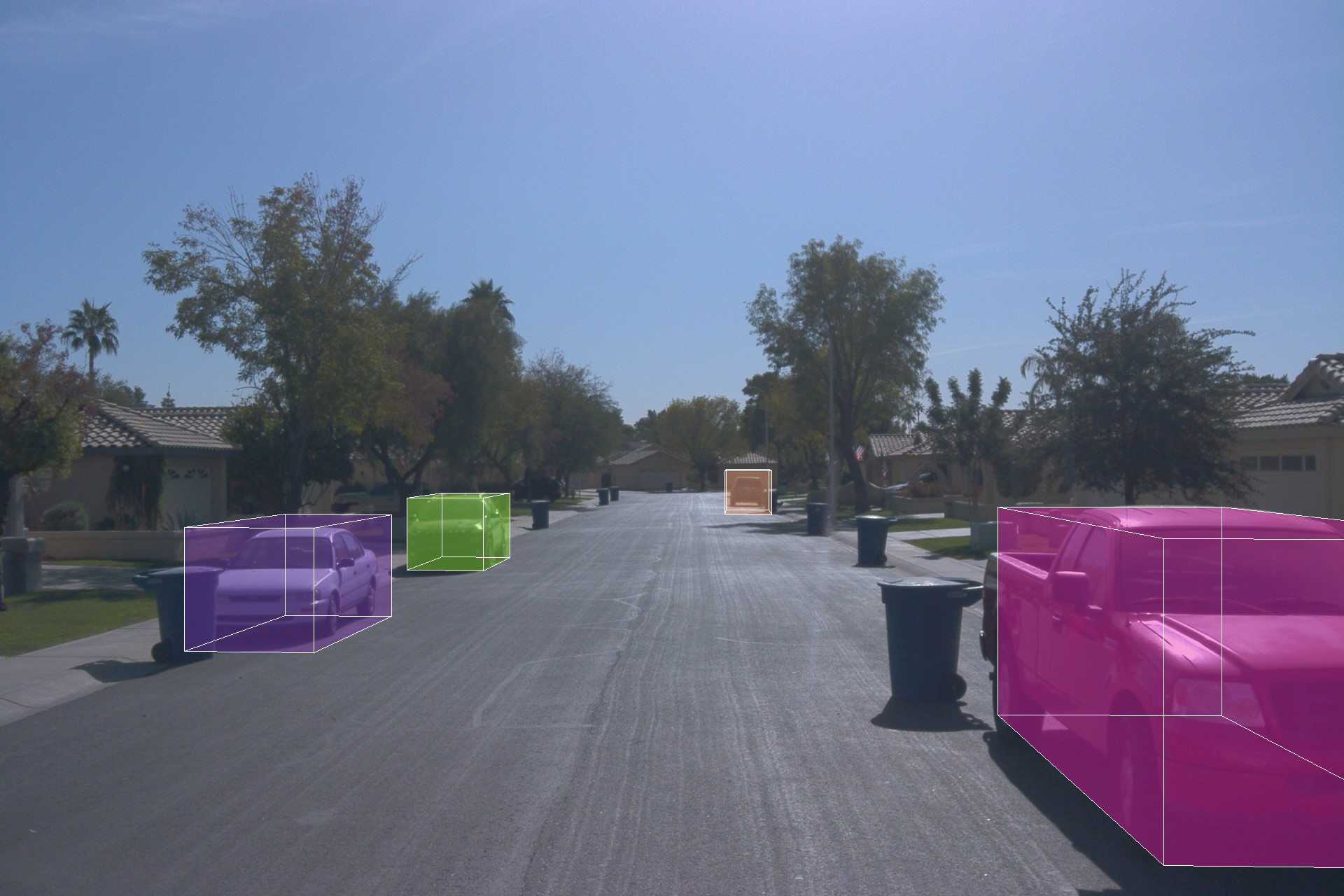} & \includegraphics[width=0.5\linewidth, trim=0px 0px 0px 100px, clip]{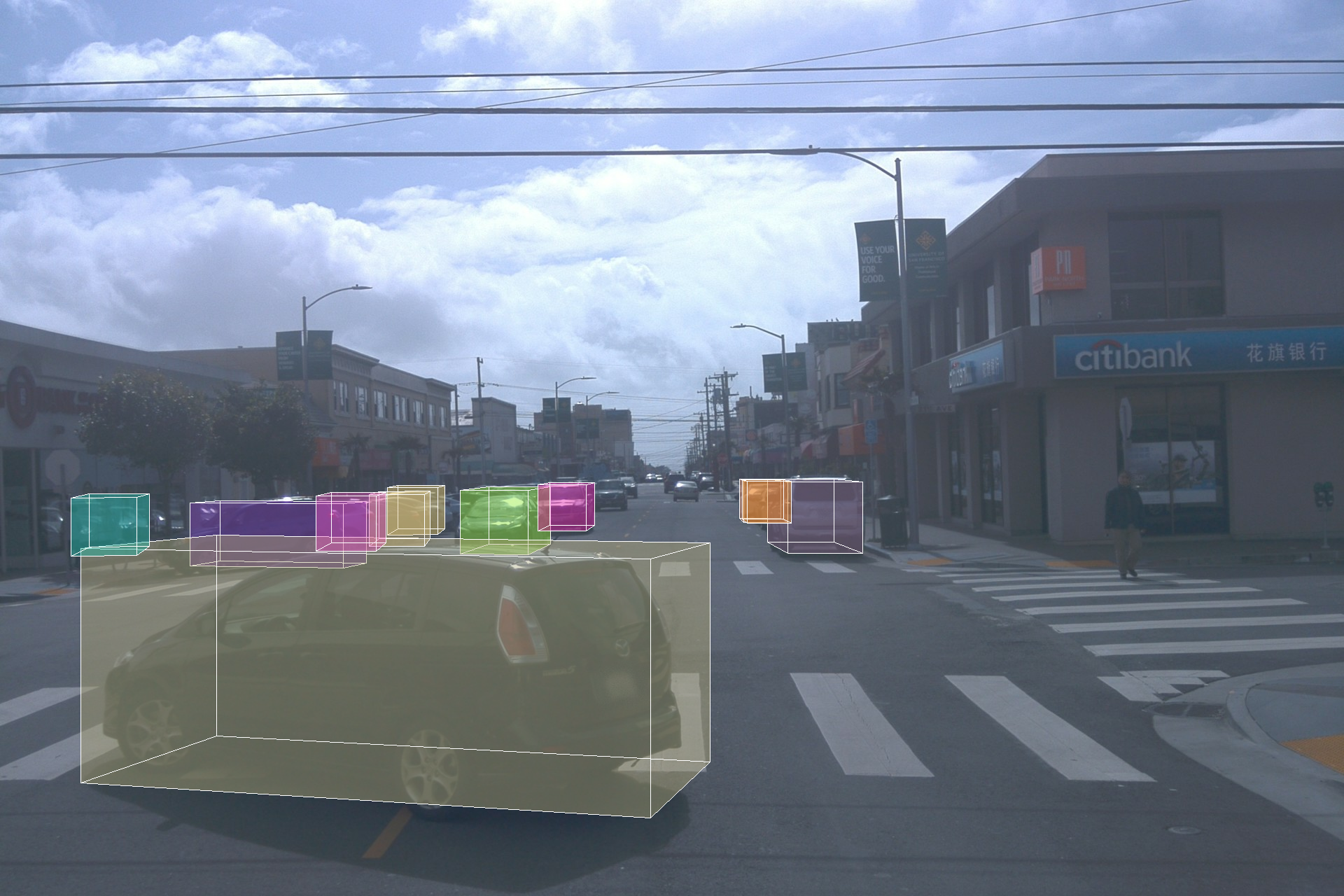} \\
    \includegraphics[width=0.5\linewidth, trim=0px 0px 0px 100px, clip]{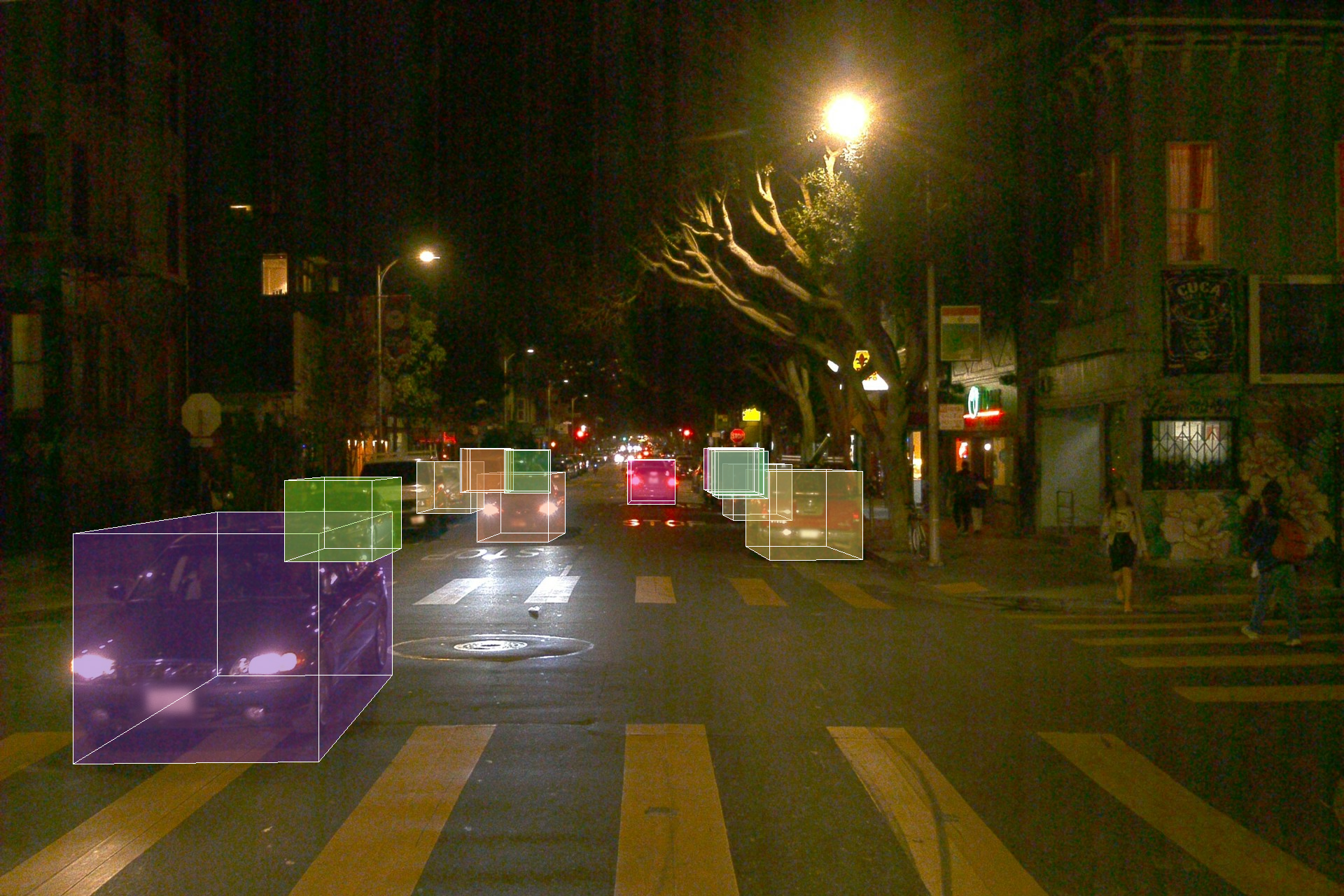} & \includegraphics[width=0.5\linewidth, trim=0px 0px 0px 100px, clip]{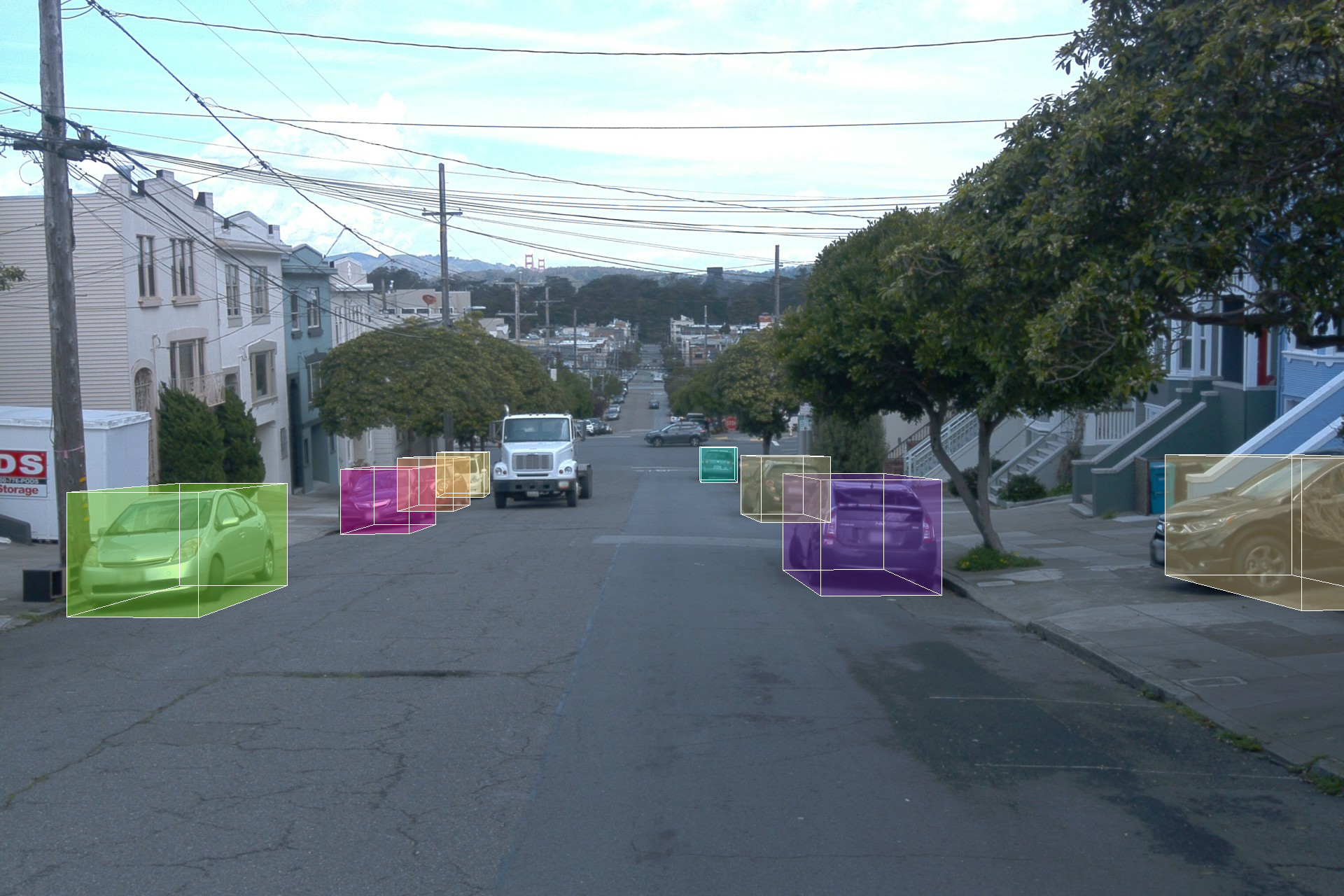} \\
\end{tabular}
\caption{Qualitative analysis of MonoDETR~\cite{zhang2023monodetr} trained using our method without using human annotations on Waymo Open Dataset~\cite{Sun_2020_CVPR}. Colourful 3D bounding boxes are predictions of our model.}
\label{fig:qualitative_analysis_supp_waymo}
\end{figure*}

In \Cref{fig:qualitative_analysis_supp_waymo}, we show how MonoDETR~\cite{zhang2023monodetr} trained by our method on Waymo Open Dataset~\cite{Sun_2020_CVPR} training set performs on the Waymo validation subset.

\clearpage
\clearpage

\end{document}